\documentclass{article} % For LaTeX2e
\usepackage{iclr2027_conference,times}

\usepackage{amsmath,amsfonts,bm}

\def\eqref#1{equation~\ref{#1}}
\def\1{\bm{1}}

\def\vmu{{\bm{\mu}}}

\def\vc{{\bm{c}}}
\def\vd{{\bm{d}}}

\def\vh{{\bm{h}}}
\def\vi{{\bm{i}}}

\def\vq{{\bm{q}}}

\def\mW{{\bm{W}}}

\DeclareMathAlphabet{\mathsfit}{\encodingdefault}{\sfdefault}{m}{sl}
\SetMathAlphabet{\mathsfit}{bold}{\encodingdefault}{\sfdefault}{bx}{n}

\usepackage{hyperref}
\usepackage{url}
\usepackage{booktabs}
\usepackage{multirow}
\usepackage{graphicx}
\usepackage{xcolor}
\usepackage{colortbl}
\usepackage{amsmath}
\usepackage{amssymb}
\usepackage{pifont}
\usepackage{tikz}
\usepackage{tcolorbox}
\usepackage{subcaption}
\usepackage{wrapfig}
\usepackage{makecell}
\usepackage{cleveref}
\tcbuselibrary{skins,breakable}
\usepackage{enumitem}

\title{From Knowing to Abstaining: \\ Bridging the Representation–Action Gap in Vision–Language Models}

\author{Jialuo He, Huangxun Chen\thanks{Corresponding author.} \\
The Hong Kong University of Science and Technology (Guangzhou) \\
\texttt{huangxunchen@hkust-gz.edu.cn} \\
}

\newcommand{\sys}{Rep2Act}

\iclrfinalcopy % Uncomment for camera-ready version, but NOT for submission.
\begin{document}

\maketitle

\begin{abstract}
The ability of VLMs to correctly abstain from answering unanswerable questions is as important as their ability to generate accurate answers to answerable ones. 
Recently, several benchmarks have emerged to evaluate and improve VLM abstention. However, they suffer from substantial limitations. First, their samples often contain unintended shortcut cues in either the images or questions that inadvertently reveal answerability; moreover, they typically provide an explicit “unanswerable” option, preventing an accurate assessment of whether VLMs can abstain spontaneously. 
Second, when used as training data, they generally provide only binary labels without fine-grained explanations to support deeper supervision.
To address these limitations, we introduce Visual Answerability Diagnosis with Rationales (VAD-R), a new benchmark constructed through a two-stage pipeline of shortcut filtering and quality verification to prevent answerability leakage. Each example is annotated with step-by-step rationales and causal evidence-gap labels. 
Evaluation of state-of-the-art open- and closed-source VLMs on VAD-R reveals strikingly limited spontaneous abstention capabilities, with the two model groups achieving average recall rates of only $11.4\%$ and $16.3\%$, respectively.
We then conduct probing analyses, revealing that hidden-state representations in certain layers can effectively distinguish answerability, yet this internal distinction fails to manifest in VLMs’ final responses.
Motivated by this observation, we introduce Rep2Act, a representation-to-action alignment method that translates latent answerability awareness into explicit abstention decisions. 
Rep2Act substantially improves action accuracy on VAD-R from $56.67\%$ to $86.33\%$ for Qwen2.5-VL-3B and from $59.33\%$ to $88.67\%$ for Qwen2.5-VL-7B.
On the out-of-distribution TUBench, Rep2Act achieves an average F1 score of $53.3\%$ with only a 3B model, surpassing the closed-source GPT-4 Turbo and GPT-4o by $16.2\%$ and $1.1\%$, respectively. 

\end{abstract}

\section{Introduction}
\label{sec:intro}

Vision-language models (VLMs)~\citep{liu2023llava,qwen25vl,dai2023instructblip,hurst2024gpt,alayrac2022flamingo,zhu2025internvl3,ye2024mplug} have achieved remarkable progress in multimodal question answering. Beyond generating correct responses to answerable questions, recent work has investigated whether and how VLMs can recognize unanswerable questions, where the requisite visual evidence is absent or otherwise insufficient, and abstain rather than fabricate answers~\citep{abstain1,abstain2,abstain3}. 
Accordingly, several abstention benchmarks have been developed~\citep{yu2024tubench,tu2023unkvqa,zheng2024mmupd,chandu2025certainlyuncertain}. However, they exhibit limitations as summarized in \Cref{tab:dataset_comparison}. 
\emph{First}, they often provide an explicit ``unanswerable'' option through binary-judgment or multiple-choice prompts. Thus, they evaluate prompted answerability judgments rather than whether models abstain spontaneously when responding to questions directly.
\emph{Second}, they typically provide only binary answerability labels without explaining why the available evidence is insufficient, thereby offering limited supervision for training models to identify the missing evidence and articulate their reasoning step by step. \emph{Third}, heuristic dataset construction can introduce unintended shortcuts. For example, TUBench~\citep{yu2024tubench} generates unanswerable tabular questions by blanking out cells, enabling an image-only probe to achieve over $99\%$ accuracy without access to the question. 
% Such scores can reflect label leakage rather than genuine evidence verification.
To systematically assess answerability-label leakage across these benchmarks, we conduct a comprehensive probing analysis. We feed each input into a frozen Qwen2.5-VL-7B-Instruct backbone and extract the hidden state of the final prompt token immediately before generation. We consider two unimodal probing conditions: \emph{Question-Only}, which includes the question text and candidate options but excludes the image, and \emph{Image-Only}, which includes the image but excludes the question text and task instructions. We train a two-layer MLP probe on the extracted features and evaluate it using five-fold stratified group cross-validation across three independent runs. As shown in Table~\ref{tab:dataset_comparison}, MM-UPD and TUBench exhibit strong unimodal predictability, indicating substantial answerability-label leakage.

\begin{table*}[t]
\small
\caption{\small Comparison of VAD-R with existing VLM unanswerability benchmarks. Text-only and image-only shortcut probes are trained on frozen Qwen2.5-VL-7B-Instruct features, with accuracy evaluated over three runs and reported as (overall average accuracy / highest subclass accuracy). } \vspace{-0.2in}
\label{tab:dataset_comparison}
\begin{center}
\resizebox{0.9\textwidth}{!}{%
\renewcommand{\arraystretch}{1.2}
\begin{tabular}{|c|c|c|c|c|c|}
\hline
\multirow{2}{*}{\textbf{Benchmark}} & \multirow{2}{*}{\makecell{\textbf{Reasoning}\\\textbf{Rationale}}} &  \multirow{2}{*}{\makecell{\textbf{Causal}\\\textbf{Label}}}  & \multirow{2}{*}{\makecell{\textbf{No Hint for}\\\textbf{Abstention}}} & \multicolumn{2}{c|}{\textbf{Shortcut Probe Diagnosis}} \\
\cline{5-6}
 & & & & \textbf{Text-only} & \textbf{Image-only} \\ 
\hline
UNK-VQA~\citep{tu2023unkvqa} & \ding{55} & \ding{55} & \ding{55} & 54.04\% / 61.46\% & 52.59\% / 58.09\%\\
\hline
MM-UPD~\citep{zheng2024mmupd} & \ding{55} & \ding{55} & \ding{55} & 90.17\% / 99.84\% & 54.37\% / 74.71\% \\
\hline
TUBench~\citep{yu2024tubench} & \ding{55} & \ding{55} & \ding{55} & 68.71\% / 77.93\% & 60.88\% / 99.67\% \\
\hline
\textbf{VAD-R (ours)}  & \ding{51} & \ding{51} (6 Categories) & \ding{51} & \textbf{48.22\% / 50\%} & \textbf{47.56\% / 48.81\%}  \\
\hline
\end{tabular}%
} \vspace{-0.2in}
\end{center}
\end{table*}

To address these limitations, we construct \textbf{Visual Answerability Diagnosis with Rationales (VAD-R)}, a benchmark based on unedited, natural photographs in which unanswerability arises from genuine gaps in visual evidence. 
To mitigate answerability-label leakage, we employ a rigorous two-stage quality-control pipeline rarely adopted by prior benchmarks: it first removes linguistically predictable shortcuts using a probe trained on frozen representations and then independently audits each question’s validity against the corresponding visual evidence. Each example is accompanied by a step-by-step rationale, while every unanswerable question is additionally assigned one or more of the six causal labels characterizing its underlying evidence gap. A stratified manual audit of 100 test examples confirms an annotation accuracy of $93\%$. 

\begin{table*}[t]
\centering
\caption{\small Summary evaluation of diverse VLMs on VAD-R test set under Direct Answering ($P_D$), where models receive only the image and question and Qwen3.6-27B serves as Action Judge, and Explicit Judgment ($P_J$), where models are explicitly prompted to judge whether the question is answerable with a binary ``yes'' or ``no''. Metrics include Action Accuracy (A-Acc), Answerable Recall (Ans-R), Abstain Recall on unanswerable queries (Abs-R), Abstain F1 (Abs-F1), and Abstain Recall Gap $\Delta\text{Abs-R} = \text{Abs-R}_{P_J} - \text{Abs-R}_{P_D}$. Action Accuracy (A-Acc) measures answer/abstain correctness from generated responses under $P_D$ and explicit yes/no judgments under $P_J$. Complete per-model evaluation results are provided in Table~\ref{tab:eight_model_disconnect_full} in Appendix~\ref{sec:appendix_full_disconnect_eval}. 
} \vspace{-0.1in}
\label{tab:eight_model_disconnect}
\resizebox{0.9\textwidth}{!}{%
\begin{tabular}{lccccccccc}
\toprule
& \multicolumn{4}{c}{\textbf{Direct Answering Mode ($P_D$)}} & \multicolumn{4}{c}{\textbf{Explicit Judgment Mode ($P_J$)}} & \textbf{Gap} \\
\cmidrule(lr){2-5}\cmidrule(lr){6-9}\cmidrule(lr){10-10}
\textbf{Model Group} & \textbf{A-Acc} & \textbf{Ans-R} & \textbf{Abs-R} & \textbf{Abs-F1} & \textbf{A-Acc} & \textbf{Ans-R} & \textbf{Abs-R} & \textbf{Abs-F1} & \textbf{$\Delta$ Abs-R} \\
\midrule
Open-Source Model (8) & 55.7 & 100.0 & 11.4 & 19.4 & 77.6 & 77.2 & 78.0 & 77.2 & \textbf{+66.6} \\
Closed-Source Model (3) & 57.2 & 98.0 & 16.3 & 27.6 & 81.7 & 68.4 & 94.9 & 84.5 & \textbf{+78.6} \\
\bottomrule
\end{tabular}%
}
\vspace{-10pt}
\end{table*}

Using VAD-R, we reveal a pronounced capability gap in VLMs between prompted answerability judgment and spontaneous abstention during direct answering. As shown in Table~\ref{tab:eight_model_disconnect}, explicit judgment yields an average abstention recall of $78.0\%$ for open-source models and $94.9\%$ for closed-source models, whereas direct answering achieves only $11.4\%$ and $16.3\%$, respectively. 
Further layerwise probing reveals that mid-to-late hidden-state representations clearly separate answerable from unanswerable queries, indicating that VLMs already encode answerability internally. Although this latent capability is elicited under explicit judgment prompting, the same signal fails to steer generation toward abstention during direct answering, i.e., \textbf{representation-to-action disconnect}. Existing
approaches~\citep{kuhn2023semantic,zhang2024vluncertainty,leng2024vcd,vardi2026clipup,zhao2024lvlmlp} do not adequately resolve this issue. As in \Cref{fig:rep2act_algorithm_overview}, they may still fabricate answers, produce generic refusals, or generate explanations unsupported by the available visual evidence.

To bridge this gap, we propose \textbf{Representation-to-Action (\sys{})}, which externalizes the model’s internal answerability signal as a discrete action $b\in\{\mathrm{A},\mathrm{U}\}$ before response generation, where A and U denote answerable and unanswerable, respectively. \sys{} combines a prior derived from deep hidden representations with vocabulary-level logits to select either an answer or abstain token. We further introduce a signed margin loss to sharpen this decision and a route-contrastive loss that assigns a higher likelihood to the ground-truth continuation under the correct route than under its inversion, thereby preventing the generated response from contradicting the selected action. 
Our main contributions are as follows:
\begin{itemize}[leftmargin=*]
    \item \textbf{Reasoning-Annotated, Leakage-Resistant Benchmark}: VAD-R provides a step-by-step rationale for every question and a six-category causal taxonomy for evidence gaps in unanswerable cases, while employing a two-stage quality-control pipeline to eliminate superficial shortcuts. 
   \item \textbf{Identification of Representation-to-Action Disconnect}: We systematically identify and characterize a fundamental disconnect in VLM abstention: answerability is encoded in deep hidden representations but fails to translate into spontaneous abstention during direct answering. 
    \item \textbf{\sys{} Framework}: We introduce \sys{}, which externalizes internal answerability representations as an explicit discrete routing variable that directly conditions subsequent response generation. \sys{} substantially improves action accuracy on VAD-R from $56.67\%$ to $86.33\%$ for Qwen2.5-VL-3B and from $59.33\%$ to $88.67\%$ for Qwen2.5-VL-7B. On the out-of-distribution TUBench benchmark, \sys{} achieves an average F1 score of $53.3\%$ with only a 3B-parameter model, outperforming GPT-4 Turbo and GPT-4o by $16.2\%$ and $1.1\%$, respectively. 
\end{itemize}

\section{VAD-R: \textbf{V}isual \textbf{A}nswerability \textbf{D}iagnosis with \textbf{R}ationales}
\label{sec:dataset}

To investigate whether VLMs genuinely understand multimodal evidence rather than exploit superficial shortcuts, we construct VAD-R as a controlled benchmark. 

\noindent\textbf{Benchmark Overview}.
\label{sec:dataset_taxonomies}
VAD-R comprises 1,852 visual question-answering instances, partitioned into 1,226 training, 326 development, and 300 test examples. 
As shown in Figure~\ref{fig:vadr_radar}(a), VAD-R covers four visual dimensions.
Every question includes a visual reasoning rationale.
For unanswerable questions, the six evidence-gap categories in Figure~\ref{fig:vadr_radar}(b) explain why the image cannot support an answer. Multiple labels may apply to a single question.
Detailed definitions and breakdowns are provided in Appendix~\ref{sec:appendix_vadr_construction}.

\noindent\textbf{Data Construction Pipeline and Quality Control}.
\label{sec:dataset_construction}
Our protocol is specifically designed to eliminate superficial shortcuts and ensure a faithful evaluation of VLMs’ genuine abstention capabilities. 
(1) We source images and descriptions from DOCCI~\citep{onoe2024docci}, whose detailed annotations capture fine-grained visual information. 
Qwen3.5-122B-A10B generates an initial pool of candidate questions across four visual dimensions, together with step-by-step visual reasoning rationales and causal evidence-gap labels. An initial quality-filtering stage yields 30,386 questions.
(2) To prevent answerability from being inferred solely from questions, we train a two-layer question-only multilayer perceptron (MLP) probe on frozen text representations from Qwen2.5-VL-7B-Instruct. We rank questions by the probe’s cross-entropy loss and remove those whose answerability labels are predictable from linguistic cues, retaining 8,357 of the most challenging questions, including 4,066 unanswerable ones.
(3) Finally, Qwen3.6-27B independently audits 4,066 candidate unanswerable questions against the corresponding visual evidence, of which 926 pass quality verification. Pairing them with an equal number of answerable examples produces the balanced benchmark of 1,852 VQA questions. Appendix~\ref{sec:appendix_vadr_construction} provides details on construction protocol and filtering statistics.

\section{In-Depth Investigation of VLM Abstention Capabilities}
\label{sec:representation_gap}

Using the genuinely challenging VAD-R benchmark, we investigate whether VLM abstention failures arise from an inability to capture visual evidence or to translate such evidence into decisions. 

\begin{figure}[t]
\centering
\begin{minipage}[t]{0.42\linewidth}
\centering
\vspace{0pt}\includegraphics[width=\linewidth]{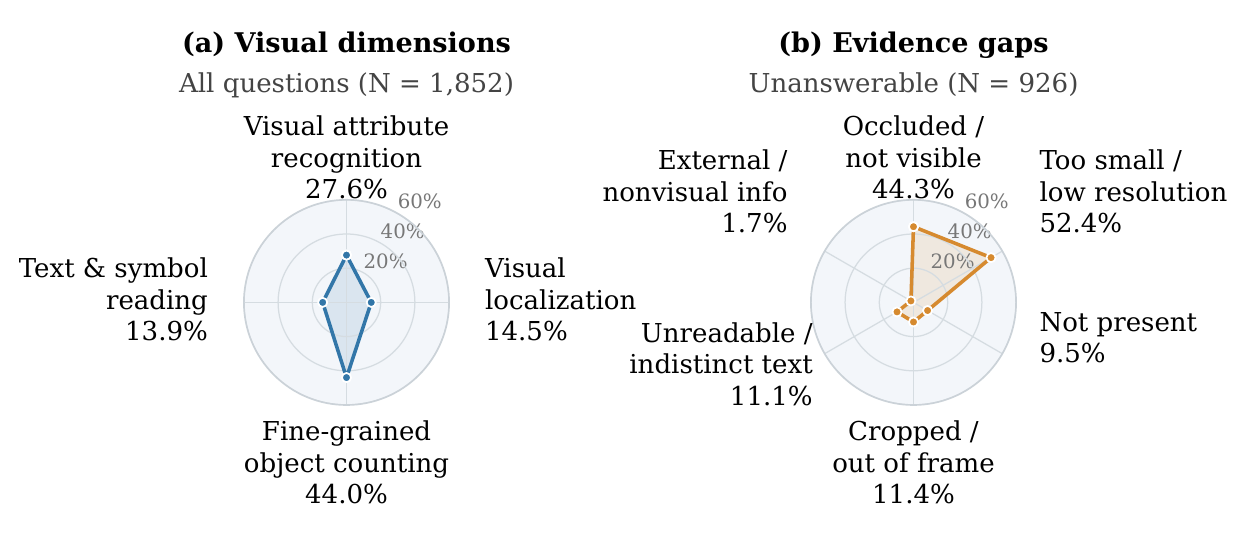}\vspace{-0.1in}
\captionsetup{font=footnotesize}
\caption{VAD-R composition across all splits. (a) Four visual dimensions among 1,852 questions. (b) Six evidence-gap categories among 926 unanswerable questions.}\vspace{-0.1in}
\label{fig:vadr_radar}
\end{minipage}\hfill
\begin{minipage}[t]{0.56\linewidth}
\centering
\vspace{0pt}\includegraphics[width=\linewidth]{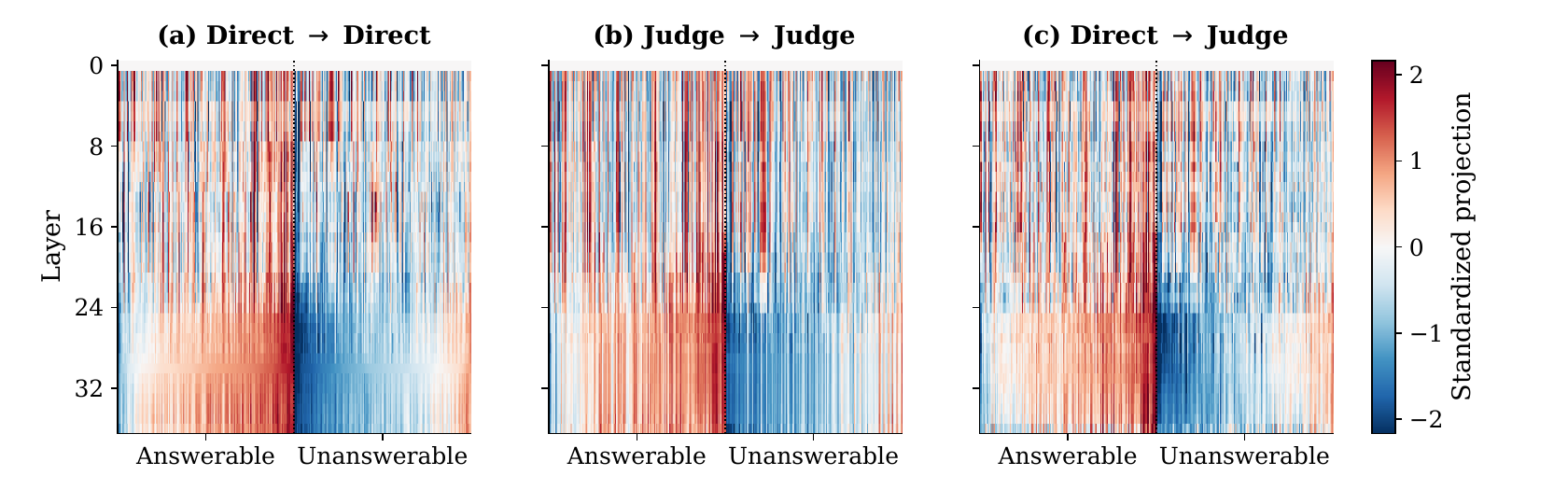}\vspace{-0.1in}
\captionsetup{font=footnotesize}
\caption{Answerability projections of Qwen2.5-VL-3B on the VAD-R development set: \textbf{(a)} Direct features on the Direct axis; \textbf{(b)} Judge features on the Judge axis; \textbf{(c)} Direct features on the Judge axis. }\vspace{-0.1in}
\label{fig:base3b_heatmap}
\end{minipage}
\end{figure}

\noindent\textbf{Examining the Separability of Answerability Representations}. 
\label{sec:direction_construction}
For an image--question pair $x=(\vi,\vq)$, answerability indicates whether the image provides sufficient evidence to answer the question, with label $y(x)\in\{\mathrm A,\mathrm U\}$. Following prior representation analyses~\citep{alain2016understanding,burns2022discover,azaria2023internal,liu2023cognitive}, we examine this signal under Direct Answering ($D$) and Explicit Judgment ($J$). At layer $l$, we define the answerability direction $\hat{\vd}_{M,l}$ as the normalized difference between the mean hidden representations of answerable and unanswerable training samples under prompting mode $M\in\{D,J\}$. The answerability score is
\begin{equation}
\label{eq:answerability_score}
z_{M,l}(x)=\frac{(\vh_l^M(x)-\vc_{M,l})^\top\hat{\vd}_{M,l}}{\sigma_{M,l}}
\end{equation}
where $\vh_l^M(x)$ is the final prompt token's hidden state, $\vc_{M,l}$ is the midpoint of the two class means, and $\sigma_{M,l}$ is the training projection scale. Positive and negative answerability scores indicate the answerable and unanswerable sides of this axis. Appendix~\ref{sec:appendix_answerability_definition} details how the answerability direction, center, and scale are computed.
As shown in \Cref{fig:base3b_heatmap}(a)(b), projecting the intermediate representations from each prompting mode onto their corresponding answerability directions reveals a clear separation between answerable and unanswerable samples in the later layers. This demonstrates that, under both prompting modes, the VLM has already encoded information relevant to determining answerability. As shown in Figure~\ref{fig:base3b_heatmap}(c), the representations remain clearly separable across prompting modes. Specifically, at layer 31 on the development split, direct-answer representations achieve an AUROC of $90.01\%$ when projected onto the judgment-mode answerability direction, compared with $91.95\%$ for judgment-mode representations projected onto their own direction, with a gap of only $1.94\%$. These results indicate that two prompting modes share a common answerability direction.

\noindent\textbf{Representation-Action Gap and Initial Attempt}.
\label{sec:feature_alignment_limit}
However, Table~\ref{tab:eight_model_disconnect} shows that the direct-answer mode achieves an average abstention recall of only $11.4\%$. 
The gap between internal answerability representations and external abstention behavior naturally raises two questions:

\noindent\textbf{Q1}: \emph{Why does VLM's internal recognition of insufficient evidence fail to manifest in its answer?} 
We hypothesize that standard VLM training overwhelmingly rewards answer generation. Instruction-tuning and preference-optimization pipelines reinforce helpful and content-producing responses~\citep{liu2023llava,qwen25vl}, biasing the autoregressive decoding distribution toward generating an answer regardless of whether the hidden states encode insufficient evidence. 
    
\noindent\textbf{Q2}: \emph{How can the answerability representation be effectively channeled into the final response?}
A natural approach is to align the direct-answer representations with the judgment-mode projections by penalizing projection discrepancies  with an auxiliary anchoring loss (Appendix~\ref{sec:appendix_centroid_geometry}). 
However, this approach yields only negligible improvements (\S\ref{sec:rq4_ablations}). The analysis in Appendix~\ref{sec:appendix_centroid_geometry} shows that the anchoring loss optimizes an internal projection score and affects the final A/U token decision only indirectly. These results suggest that regularizing continuous latent representations alone is insufficient to overcome the model's generative bias toward answering. We therefore introduce \sys{} to directly optimize the A/U decision boundary and train subsequent generation to follow the selected route.

\section{Rep2Act: \textbf{Rep}resentation-to-\textbf{Act}ion Alignment Framework}
\label{sec:rep2act}

We propose \textbf{Representation-to-Action (Rep2Act)}, a method that translates internal answerability representations into explicit answer-or-abstain decisions that condition subsequent response. 

\begin{figure*}[t]
    \centering
    \includegraphics[width=\linewidth]{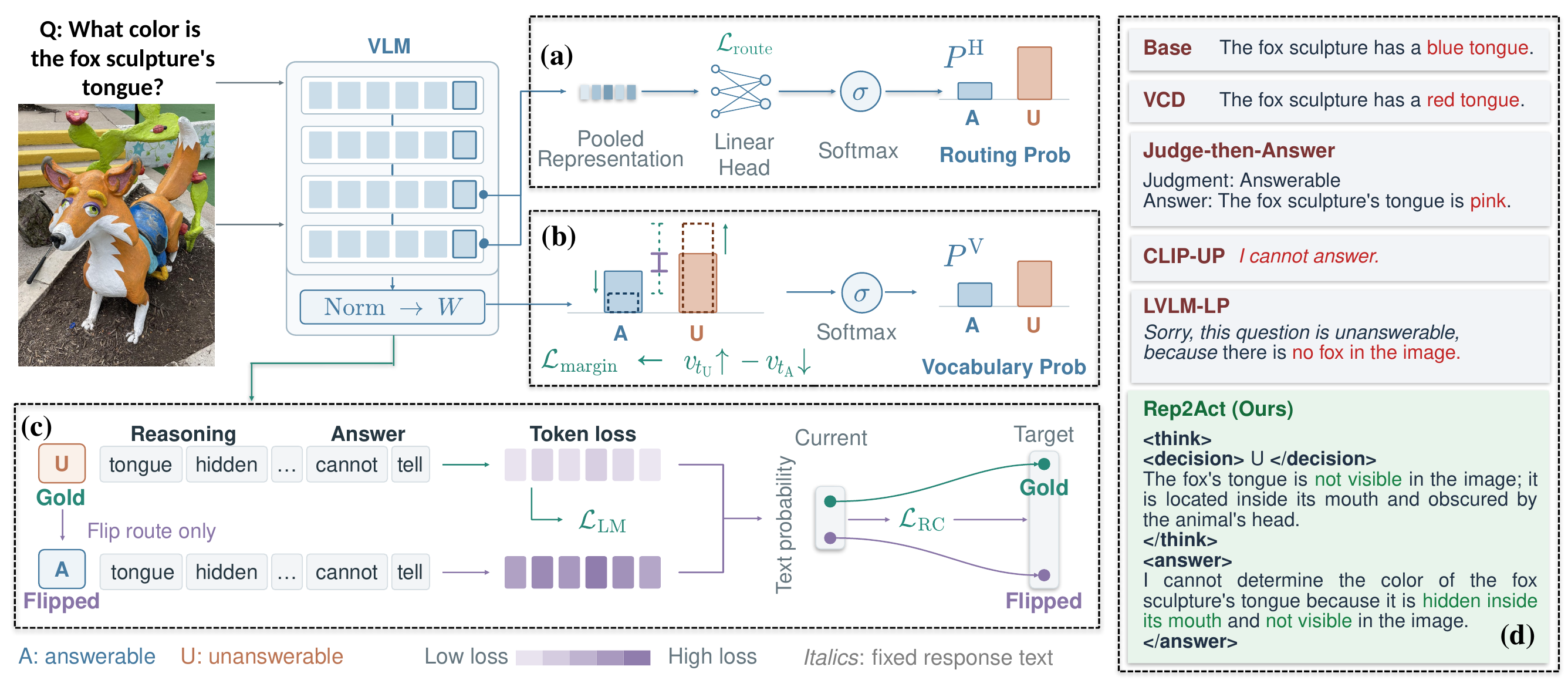}
    \caption{Rep2Act overview. (a) Representation-based routing with $\mathcal L_{\mathrm{route}}$. (b) A/U vocabulary logits trained with Signed Margin Loss. (c) Language modeling and Route-Contrastive Loss train the response to follow the route. (d) A VAD-R example comparing six 3B methods.}\vspace{-0.1in}
    \label{fig:rep2act_algorithm_overview}
\end{figure*}

As illustrated in \Cref{fig:rep2act_algorithm_overview}(d), rather than relying on unguided direct-answer generation, \sys{} structures the VLM output as a triple $(b,r,a)$ and autoregressively factorizes the corresponding generative probability as follows: 
\begin{equation}
\label{eq:rep2act_factorization}
\pi_\theta(b, r, a \mid \vi, \vq) = \underbrace{\pi_\theta(b \mid \vi, \vq)}_{\text{Discrete Routing Token}} \cdot \underbrace{\pi_\theta(r \mid b, \vi, \vq)}_{\text{Route-conditioned Rationale}} \cdot \underbrace{\pi_\theta(a \mid b, r, \vi, \vq)}_{\text{Grounded Answer Response}}
\end{equation}
Here, $\pi_\theta$ denotes the VLM parameterized by $\theta$, and $b\in\{\mathrm{A},\mathrm{U}\}$ is the routing token that specifies whether the model should answer or abstain, whereas $y(x)$ denotes the ground-truth answerability label of sample $x$. The routing token is label-consistent when $b=y(x)$ and label-contradictory when $b\neq y(x)$. Furthermore, $r=(w_1^r,\ldots,w_{|r|}^r)$ denotes the reasoning trace, and $a$ denotes the final response. Together, these components form the structured output: 
$\langle\mathrm{think}\rangle
\langle\mathrm{decision}\rangle b\langle/\mathrm{decision}\rangle
r
\langle/\mathrm{think}\rangle
\langle\mathrm{answer}\rangle a\langle/\mathrm{answer}\rangle$.

\noindent\textbf{\sys{} Training Scheme}. An overview of the \sys{} training pipeline is presented in \Cref{fig:rep2act_algorithm_overview}. Specifically, we introduce a set of complementary supervision signals that systematically guide the VLM in translating internal answerability representations into explicit answer-or-abstain decisions. 

$\bullet$~\emph{Routing-token Classification Loss}.
\label{sec:route_head}
Guided by the layerwise probing analysis in Appendix~\ref{sec:appendix_layer_selection}, we select the layers $\mathcal K_{\mathrm{H}}$ at which answerable and unanswerable representations are most clearly separable and aggregate their hidden states at the final prompt token as follows: 
\begin{equation}
\label{eq:route_head_pooling}
\bar{\vh}_{\text{route}} = \frac{1}{|\mathcal K_{\mathrm{H}}|} \sum_{l \in \mathcal K_{\mathrm{H}}} \frac{\vh_l^{D}(x)}{\|\vh_l^{D}(x)\|_2} \in \mathbb{R}^d
\end{equation}
We apply layer normalization (L-N) to the pooled representation $\bar{\vh}_{\mathrm{route}}$, followed by a linear classification head parameterized by the weight matrix $\mW_h\in\mathbb{R}^{2\times d}$ and bias vector $\boldsymbol{\beta}_h\in\mathbb{R}^2$, to obtain the routing logits $(\eta_{\mathrm{A}},\eta_{\mathrm{U}})$ and their corresponding probabilities (see \Cref{fig:rep2act_algorithm_overview} (a)): 
\[
(\eta_{\text{A}}, \eta_{\text{U}}) = \mW_h\operatorname{L-N}(\bar{\vh}_{\text{route}})+\boldsymbol\beta_h, 
P^{\mathrm H}(\text{A}\mid\vi,\vq) =\frac{e^{\eta_{\text{A}}}}{e^{\eta_{\text{A}}}+e^{\eta_{\text{U}}}},
P^{\mathrm H}(\text{U}\mid\vi,\vq)=1-P^{\mathrm H}(\text{A}\mid\vi,\vq)
\]
We then define the routing-token classification loss for each sample as follows:
\begin{equation}
\label{eq:loss_route}
\mathcal{L}_{\text{route}}(x) = - \mathbb{I}(y = \text{A}) \log P^{\mathrm{H}}(\text{A} \mid \vi, \vq) - \mathbb{I}(y = \text{U}) \log P^{\mathrm{H}}(\text{U} \mid \vi, \vq)
\end{equation}
where $\mathbb{I}$ denotes the indicator function, which equals 1 when its condition holds and 0 otherwise. 

$\bullet$~\emph{Vocabulary Route Logits and Signed Margin Loss}.
\label{sec:margin_loss}
As shown in \Cref{fig:rep2act_algorithm_overview}(b), we further make the routing variable explicit. When predicting the A/U token following $\langle\text{think}\rangle\langle\text{decision}\rangle$, the VLM produces a vocabulary-logit vector $\mathbf{v}\in\mathbb{R}^{|\mathcal{V}|}$, where each entry is an unnormalized score for a token in the vocabulary $\mathcal{V}$. 
We extract $v_{t_{\text{A}}}=\mathbf{v}[t_{\text{A}}]$ and $v_{t_{\text{U}}}=\mathbf{v}[t_{\text{U}}]$, where $t_{\text{A}}$ and $t_{\text{U}}$ denote the vocabulary indices of the A and U tokens, respectively. We then apply a softmax over these two logits to obtain the answerability probabilities. 
\begin{equation}
\label{eq:vocab_prob}
P^{\mathrm{V}}(\text{A} \mid \vi, \vq) = \frac{\exp(v_{t_{\text{A}}})}{\exp(v_{t_{\text{A}}}) + \exp(v_{t_{\text{U}}})}, \quad P^{\mathrm{V}}(\text{U} \mid \vi, \vq) = \frac{\exp(v_{t_{\text{U}}})}{\exp(v_{t_{\text{A}}}) + \exp(v_{t_{\text{U}}})}
\end{equation}
To promote robust routing decisions and penalize ambiguous predictions, we introduce a signed margin loss as follows: 
\begin{equation}
\label{eq:loss_margin}
\mathcal{L}_{\text{margin}}(\theta) = \max\left(0, \; \gamma - s \cdot \left(v_{t_{\text{A}}} - v_{t_{\text{U}}}\right)\right)
\end{equation}
where $s\in\{+1,-1\}$ denotes the ground-truth direction, with $s=+1$ if $y=\text{A}$ and $s=-1$ if $y=\text{U}$, and $\gamma>0$ is the target margin. As shown in \Cref{eq:loss_margin}, achieving zero loss requires $v_{t_{\text{A}}}-v_{t_{\text{U}}}\geq\gamma$ for answerable inputs and $v_{t_{\text{U}}}-v_{t_{\text{A}}}\geq\gamma$ for unanswerable inputs, thereby encouraging a wider decision margin. 

$\bullet$~\emph{Route-Contrastive Loss}.
\label{sec:control_loss}
Even when the generated routing token matches the ground-truth label, the subsequent generation may contradict the routing decision, a phenomenon we term \emph{route drift}. To address this issue, \sys{} introduces a route-contrastive loss that encourages the continuation to remain consistent with the selected route, as illustrated in \Cref{fig:rep2act_algorithm_overview}(c). To compute the route-contrastive loss, we perform two teacher-forced forward passes, with the routing token set to $b=y(x)$ and $b=\bar y(x)$, respectively. Only the routing token changes; the target reasoning and answer remain unchanged. We compute $\mathcal L_{\mathrm{LM}}$ from the gold-route pass and use the continuation likelihoods from both passes to compute $\mathcal L_{\mathrm{RC}}$.
To mitigate the effect of sequence length, for each component $k\in\{r,a\}$, corresponding to the reasoning process and the final answer, respectively, we compute the mean negative log-likelihood (NLL) separately: 
\begin{equation}
\label{eq:route_span_nll}
N_k(b)=-\frac{1}{|\mathcal T_k|}\sum_{t\in\mathcal T_k}
\log\pi_\theta(o_t\mid\mathbf o_{<t}^{b},\vi,\vq)
\end{equation}
where $\mathcal{T}_k$ denotes the set of token positions corresponding to component $k$ in the gold output sequence $\mathbf{o}$. The context $\mathbf{o}_{<t}^{b}$ contains the selected routing token $b\in\{y,\bar{y}\}$ and all preceding gold tokens. 
Then, for each component, we subtract the gold-route NLL, $N_k(y)$, from the flipped-route NLL, $N_k(\bar{y})$, and then average the resulting differences:  
\begin{equation}
\label{eq:route_contrast_gap}
g=\frac{1}{2}\underbrace{\left[N_r(\bar y)-N_r(y)\right]}_{\text{Reasoning NLL difference}}
+\frac{1}{2}\underbrace{\left[N_a(\bar y)-N_a(y)\right]}_{\text{Answer NLL difference}}
\end{equation}
A positive $g$ indicates that, on average, the gold reasoning process and answer are easier to generate under the gold route than under the flipped route. 
We then define the route-contrastive loss as
\begin{equation}
\label{eq:loss_control}
\mathcal{L}_{\mathrm{RC}}(\theta)=\max(0,\delta-g)
\end{equation}
where $\delta>0$ is the target margin. This objective encourages the difference $g$ to reach $\delta$. 

\emph{Putting It All Together.}
Overall, \sys{} combines the language-modeling, route-classification, signed-margin, and route-contrastive objectives as follows: 
\begin{equation}
\label{eq:total_loss}
\mathcal{L}_{\text{total}} = \mathcal{L}_{\text{LM}} + \lambda_r \mathcal{L}_{\text{route}} + \lambda_m \mathcal{L}_{\text{margin}} + \lambda_c \mathcal{L}_{\mathrm{RC}}
\end{equation}

The training of \sys{} proceeds in two stages. \emph{Stage 1: Routing-head training.} We freeze the VLM backbone and optimize only the representation-based routing head on VAD-R using $\mathcal{L}_{\mathrm{route}}$. \emph{Stage 2: Joint training.} We initialize the routing head with the parameters learned in Stage 1 and jointly optimize it with the backbone’s LoRA attention adapters using $\mathcal{L}_{\mathrm{total}}$.

\noindent\textbf{\sys{} Test-Time Inference Scheme}.
\label{sec:inference_fusion}
At inference time, \sys{} combines the continuous prior derived from the deep representation, $P^{\mathrm{H}}$ with the vocabulary-level probability, $P^{\mathrm{V}}$:
\begin{equation}
\label{eq:ensemble_prob}
P^{\mathrm{E}}(\text{A}\mid\vi,\vq)
= \alpha P^{\mathrm{H}}(\text{A}\mid\vi,\vq) + (1-\alpha)P^{\mathrm{V}}(\text{A}\mid\vi,\vq)
\end{equation}
where $\alpha\in[0,1]$ balances the representation signal against the output-layer confidence. 
We obtain the discrete routing token $\hat{b}$ by thresholding the fused probability: $\hat{b}=\text{A}$ if $P^{\mathrm{E}}(\text{A}\mid\vi,\vq)\geq 0.5$, and $\hat{b}=\text{U}$ otherwise. We then insert $\hat{b}$ into the decision tag and generate the reasoning $\hat{r}$ and final answer $\hat{a}$ conditioned on the routing token, image, and question: $(\hat{r},\hat{a})\sim\pi_\theta(\cdot\mid\hat{b},\vi,\vq)$.

\section{Evaluation}
\label{sec:experiments}

We evaluate \sys{} by addressing the following research questions:
\textbf{RQ1: In-Domain Effectiveness and Preservation of Standard VQA Capabilities.} How effectively does \sys{} bridge the representation-to-action disconnect compared with existing baselines? How do its improvements vary across different visual skills? Can \sys{} improve answerability discrimination without degrading performance on standard VQA tasks? 
\textbf{RQ2: Out-of-Distribution Generalization.} How robustly does \sys{} generalize to previously unseen domains and types of unanswerability? 
% (3) \textbf{RQ3: Preservation of Standard VQA Capabilities.} 
\textbf{RQ3: Ablation Study.} What are the respective contributions of the signed margin loss, route-contrastive loss, and supervised target formulations? 

% \subsection{Experimental Setup}
% \label{sec:experimental_setup}

\begin{table*}[t]
\caption{Results on VAD-R test set. Rep2Act, Rep2Act$^{\dagger}$, and Rep2Act$^{\ddagger}$ denote the fused-decision variant ($\alpha=0.50$), routing-head-only variant ($\alpha=1.0$), and vocabulary-only variant ($\alpha=0.0$).} \vspace{-0.1in}
\label{tab:unified_id_results}
\label{tab:unified_vad_r_only_results}
\centering
\begin{subtable}[t]{0.505\textwidth}
\caption{Answerability classification.}\vspace{-0.05in}
\label{tab:unified_id_decision}
\centering
\setlength{\tabcolsep}{3.5pt}
\scriptsize
\resizebox{0.9\linewidth}{!}{%
\begin{tabular}{lccccc}
\toprule
\textbf{Method} & \textbf{Acc} & \textbf{M-P} & \textbf{M-R} & \textbf{M-F1} & \textbf{AUROC} \\
\midrule
\multicolumn{6}{l}{\textit{Qwen2.5-VL-3B}} \\
SE & 69.00 & 69.25 & 69.00 & 68.90 & 69.44 \\
VCD & 50.33 & 50.34 & 50.33 & 50.13 & 53.39 \\
VL-U & 58.67 & 62.11 & 58.67 & 55.50 & 61.01 \\
LVLM-LP & 80.67 & 82.13 & 80.67 & 80.44 & 90.71 \\
Rep2Act$^{\dagger}$ & 82.33 & 83.59 & 82.33 & 82.17 & 92.60 \\
Rep2Act$^{\ddagger}$ & \underline{86.33} & \underline{86.61} & \underline{86.33} & \underline{86.31} & \underline{92.68} \\
Rep2Act & \textbf{87.33} & \textbf{87.33} & \textbf{87.33} & \textbf{87.33} & \textbf{92.98} \\
\midrule
\multicolumn{6}{l}{\textit{Qwen2.5-VL-7B}} \\
SE & 67.67 & 68.26 & 67.67 & 67.40 & 69.30 \\
VCD & 46.33 & 46.19 & 46.33 & 45.83 & 46.17 \\
VL-U & 58.67 & 59.96 & 58.67 & 57.28 & 58.87 \\
LVLM-LP & \underline{87.33} & 88.01 & \underline{87.33} & 87.28 & \textbf{95.82} \\
Rep2Act$^{\dagger}$ & 63.33 & 78.85 & 63.33 & 57.64 & 91.84 \\
Rep2Act$^{\ddagger}$ & \textbf{88.33} & \underline{88.42} & \textbf{88.33} & \textbf{88.33} & 93.75 \\
Rep2Act & \textbf{88.33} & \textbf{88.54} & \textbf{88.33} & \underline{88.32} & \underline{93.82} \\
\bottomrule
\end{tabular}%
}
\end{subtable}%
\hfill
\begin{subtable}[t]{0.45\textwidth}
\caption{Response generation and content.}\vspace{-0.05in}
\label{tab:unified_id_action}
\centering
\setlength{\tabcolsep}{3.5pt}
\scriptsize
\resizebox{0.9\linewidth}{!}{%
\begin{tabular}{lccccc}
\toprule
\textbf{Method} & \textbf{A-Acc} & \textbf{M-P} & \textbf{M-R} & \textbf{M-F1} & \textbf{C-Acc} \\
\midrule
\multicolumn{6}{l}{\textit{Qwen2.5-VL-3B}} \\
$P_D$ & 56.67 & 76.79 & 56.67 & 46.65 & 30.00 \\
$P_{JA}$ & 72.00 & 76.53 & 72.00 & 70.75 & 32.33 \\
VCD & 58.33 & 75.44 & 58.33 & 49.91 & 30.33 \\
CLIP-UP & 63.33 & 76.63 & 63.33 & 58.10 & 30.00 \\
LVLM-LP & 81.33 & 82.64 & 81.33 & 81.14 & 26.00 \\
Rep2Act$^{\dagger}$ & 81.67 & 81.91 & 81.67 & 81.63 & 53.67 \\
Rep2Act$^{\ddagger}$ & \underline{84.00} & \underline{84.62} & \underline{84.00} & \underline{83.93} & \underline{55.67} \\
Rep2Act & \textbf{86.33} & \textbf{86.33} & \textbf{86.33} & \textbf{86.33} & \textbf{57.00} \\
\midrule
\multicolumn{6}{l}{\textit{Qwen2.5-VL-7B}} \\
$P_D$ & 59.33 & 77.57 & 59.33 & 51.28 & 43.67 \\
$P_{JA}$ & 85.00 & 85.04 & 85.00 & 85.00 & 47.00 \\
VCD & 60.33 & 76.39 & 60.33 & 53.22 & 42.00 \\
CLIP-UP & 61.67 & 71.34 & 61.67 & 56.77 & 29.00 \\
LVLM-LP & \underline{87.00} & 87.74 & \underline{87.00} & 86.94 & 47.33 \\
Rep2Act$^{\dagger}$ & 68.67 & 77.20 & 68.67 & 66.00 & \underline{48.33} \\
Rep2Act$^{\ddagger}$ & \textbf{88.67} & \textbf{88.69} & \textbf{88.67} & \underline{88.66} & \textbf{60.67} \\
Rep2Act & \textbf{88.67} & \underline{88.67} & \textbf{88.67} & \textbf{88.67} & \textbf{60.67} \\
\bottomrule
\end{tabular}%
}
\end{subtable}
\vspace{-5pt}
\end{table*}

\begin{figure}[t]
\centering
\includegraphics[width=\textwidth]{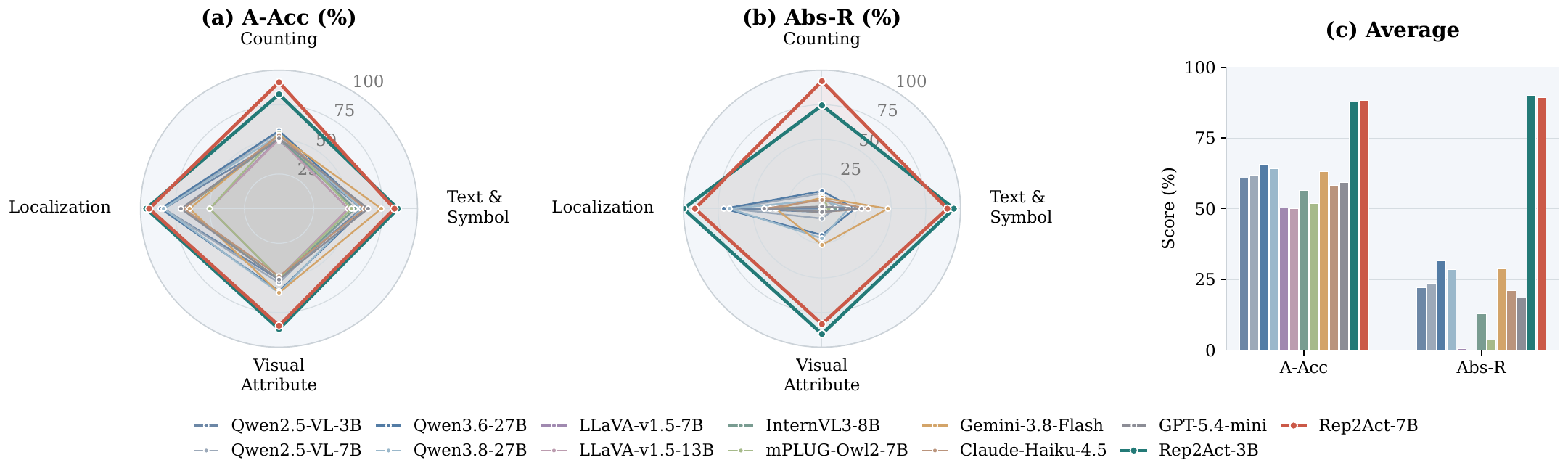}\vspace{-0.1in}
\caption{Direct Answering ($P_D$): four-skill radar plots of (a) action accuracy (A-Acc) and (b) abstention recall (Abs-R), with (c) their averages across 4 visual skills.}\vspace{-0.1in}
\label{fig:skill_direct_radar}
\vspace{-5pt}
\end{figure}

\noindent\textbf{Evaluation Benchmarks.}
We evaluate \sys{} on \emph{VAD-R} for in-domain answerability diagnosis; \emph{TUBench}~\citep{yu2024tubench} and \emph{UNK-VQA}~\citep{tu2023unkvqa} for out-of-distribution generalization; and \emph{POPE}~\citep{li2023pope}, \emph{GQA}~\citep{hudson2019gqa}, and \emph{TextVQA}~\citep{singh2019textvqa} for general multimodal perception. 
% These benchmarks span diverse domains, including code, mathematics, scene text, spatial reasoning, and visual robustness. 
Complete benchmark descriptions are provided in Appendix~\ref{sec:appendix_datasets}. 

\noindent\textbf{Baselines.}
We compare \sys{} against the following baselines. \emph{Semantic Entropy (SE)}~\citep{kuhn2023semantic} and \emph{VL-Uncertainty (VL-U)}~\citep{zhang2024vluncertainty} produce answerability classifications only. \emph{Visual Contrastive Decoding (VCD)}~\citep{leng2024vcd}, \emph{LVLM-LP}~\citep{zhao2024lvlmlp} support both answerability classification and complete response generation. \emph{CLIP-UP}~\citep{vardi2026clipup}, \emph{Direct Answering} ($P_D$), and \emph{Judge-then-Answer} ($P_{JA}$) are evaluated based on their complete responses, which either provide an answer or abstain. Implementation details are in Appendix~\ref{sec:appendix_baselines}. 

\noindent\textbf{Evaluation Metrics.}
We evaluate answerability classification against the ground-truth labels using accuracy (Acc), macro-precision (M-P), macro-recall (M-R), macro-F1 (M-F1), and, for continuous uncertainty scores, the area under the receiver operating characteristic curve (AUROC). For complete generated responses, we measure action accuracy (A-Acc), using Qwen3.6-27B as an action judge to determine whether each response answers or abstains. On VAD-R, we additionally evaluate content accuracy (C-Acc) using Qwen3.8-27B, which requires factually correct responses to answerable questions and valid causal explanations of the missing visual evidence for unanswerable questions. All metrics are reported as percentages. The evaluation protocols and judge prompts are detailed in Appendix~\ref{sec:appendix_training_details} and Appendix~\ref{sec:appendix_prompts_judges}, respectively. 

\begin{table*}[t]
\caption{OOD evaluation on TUBench and UNK-VQA.}\vspace{-0.1in}
\label{tab:unified_ood_combined}
\centering
\begin{subtable}[t]{0.485\textwidth}
\caption{Answerability classification.}\vspace{-0.05in}
\label{tab:combined_ood_decision}
\centering
\setlength{\tabcolsep}{4.2pt}
\scriptsize
\resizebox{0.9\linewidth}{!}{%
\begin{tabular}{lccccc}
\toprule
\textbf{Method} & \textbf{Acc} & \textbf{M-P} & \textbf{M-R} & \textbf{M-F1} & \textbf{AUROC} \\
\midrule
\multicolumn{6}{l}{\textit{TUBench}} \\
SE & \underline{52.72} & 53.63 & \underline{53.13} & \underline{51.23} & 53.54 \\
VCD & 47.54 & 46.90 & 48.27 & 41.48 & 52.11 \\
VL-U & 51.10 & \underline{53.67} & 51.88 & 44.77 & 53.12 \\
LVLM-LP & 50.30 & 50.27 & 50.27 & 50.27 & \underline{57.07} \\
Rep2Act & \textbf{54.21} & \textbf{54.17} & \textbf{54.16} & \textbf{54.16} & \textbf{57.19} \\
\midrule
\multicolumn{6}{l}{\textit{UNK-VQA}} \\
SE & 52.26 & 50.68 & 50.54 & \textbf{48.73} & 51.34 \\
VCD & 47.12 & 48.48 & 48.74 & 45.47 & 47.97 \\
VL-U & 53.83 & 52.11 & 50.70 & 42.29 & 50.96 \\
LVLM-LP & \underline{54.49} & \underline{56.49} & 50.95 & 39.26 & \underline{55.55} \\
Rep2Act & \textbf{55.37} & \textbf{57.23} & \textbf{52.17} & \underline{43.58} & \textbf{57.71} \\
\bottomrule
\end{tabular}%
}
\end{subtable}%
\hspace{0.1in}
\begin{subtable}[t]{0.4\textwidth}
\caption{Response-level actions.}\vspace{-0.05in}
\label{tab:combined_ood_action}
\centering
\setlength{\tabcolsep}{4.2pt}
\scriptsize
\resizebox{0.9\linewidth}{!}{%
\begin{tabular}{lcccc}
\toprule
\textbf{Method} & \textbf{A-Acc} & \textbf{M-P} & \textbf{M-R} & \textbf{M-F1} \\
\midrule
\multicolumn{5}{l}{\textit{TUBench}} \\
Base-$P_D$ & 53.27 & 59.29 & 52.31 & 41.77 \\
VCD & \underline{55.65} & \underline{61.31} & \underline{54.82} & 47.70 \\
CLIP-UP & 55.35 & 59.73 & 54.55 & 48.00 \\
LVLM-LP & 53.31 & 53.22 & 53.12 & \textbf{52.86} \\
Rep2Act & \textbf{56.84} & \textbf{61.41} & \textbf{56.09} & \underline{50.71} \\
\midrule
\multicolumn{5}{l}{\textit{UNK-VQA}} \\
Base-$P_D$ & 50.64 & \underline{55.89} & \underline{53.24} & 45.83 \\
VCD & 52.43 & 52.18 & 52.18 & \textbf{52.18} \\
CLIP-UP & 51.17 & 52.84 & 52.51 & 50.26 \\
LVLM-LP & \underline{54.69} & \textbf{56.65} & 51.24 & 40.38 \\
Rep2Act & \textbf{55.38} & 54.62 & \textbf{53.47} & \underline{51.26} \\
\bottomrule
\end{tabular}%
}
\end{subtable}
\end{table*}

\begin{figure}[t]
\centering\vspace{-0.1in}
\includegraphics[width=\textwidth]{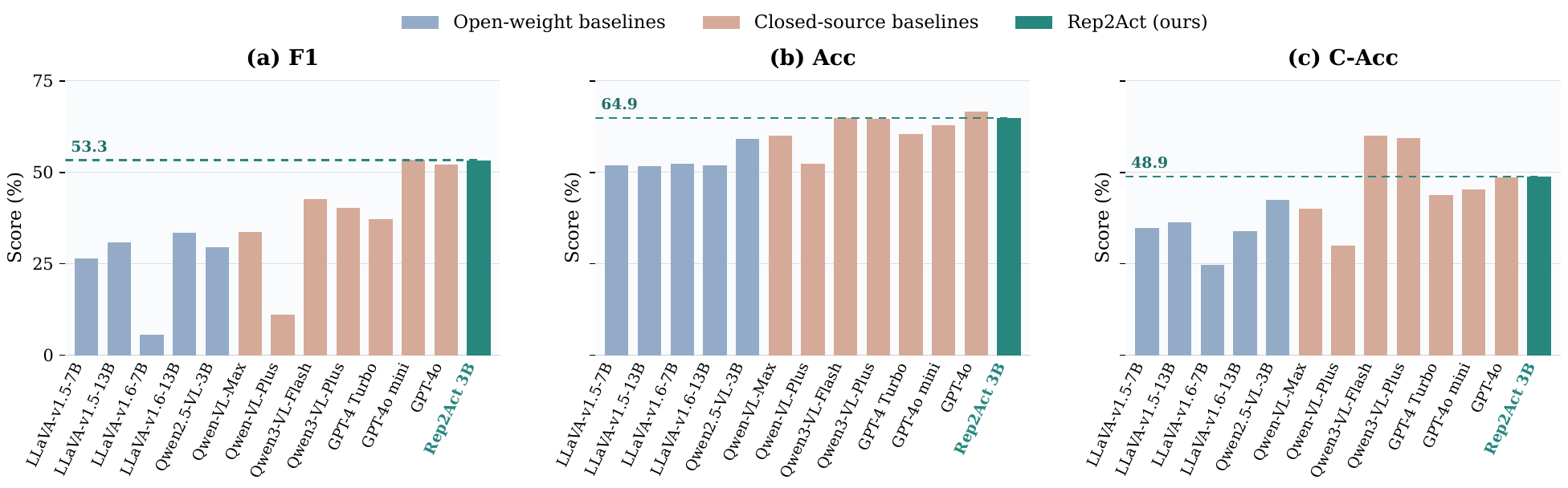}\vspace{-0.1in}
\caption{Results under the official TUBench evaluation protocol, averaged across four subsets: (a) F1, (b) accuracy (Acc), and (c) content accuracy (C-Acc).}
\label{fig:tubench_official_radar}
\vspace{-5pt}
\end{figure}

\subsection{In-Domain Effectiveness and Baseline Comparison (RQ1)}
\label{sec:rq1_in_domain}

\noindent\textbf{In-Domain Results on VAD-R.}
\label{sec:in_domain}
As shown in Table~\ref{tab:unified_id_results}(a), for 3B model, the standalone representation-based routing head, Rep2Act$^{\dagger}$ with $\alpha=1.0$, achieves an accuracy of $82.33\%$. Fusing its predictions with the vocabulary logits using $\alpha=0.50$ increases accuracy to $87.33\%$, outperforming the vocabulary-only routing variant, Rep2Act$^{\ddagger}$, which achieves $86.33\%$. 
For 7B model, vocabulary-based routing alone already achieves $88.33\%$ accuracy, matching the performance of the fused variant.
Table~\ref{tab:unified_id_results}(b) shows that LVLM-LP achieves $81.33\%$ A-Acc but only $26.00\%$ C-Acc with 3B model, falling below the base model’s $30.00\%$. 
For inputs predicted to be unanswerable, LVLM-LP inserts the fixed prefix ``Sorry, this question is unanswerable, because'' and relies on the frozen 3B base model to generate the continuation. Consequently, the model may provide an incorrect justification for abstention, as illustrated in Figure~\ref{fig:rep2act_algorithm_overview}(d). More examples are in Appendix~\ref{sec:additional_response_cases}. 

% shows that representation-based routing provides vital auxiliary decision signals.

\noindent\textbf{Fine-Grained Visual Skill Breakdown.}
\label{sec:rq1_skill_breakdown}
Figure~\ref{fig:skill_direct_radar} shows that baseline performance varies substantially across visual skills, particularly in abstention recall. In contrast, both Rep2Act models achieve consistently high A-Acc and Abs-R across all four skills, demonstrating robust improvements across diverse dimensions of visual understanding. Detailed comparisons across visual skills and prompting modes are provided in Appendix~\ref{sec:appendix_full_disconnect_eval}, particularly Table~\ref{tab:skill_breakdown}.

\noindent\textbf{Preservation of Standard VQA Capabilities.}
\label{sec:rq3_vqa_preservation}
Table~\ref{tab:general_vqa_results} evaluates capability preservation on POPE, GQA, and TextVQA. Rep2Act preserves the general multimodal perception capabilities of base models, with performance deviations of no more than $0.38\%$ for 3B model and $0.64\%$ for 7B model. 

\subsection{Out-of-Distribution Generalization (RQ2)}
\label{sec:rq2_ood}
To assess whether representation-to-action routing generalizes beyond the in-domain distribution, we evaluate Rep2Act on TUBench~\citep{yu2024tubench} and UNK-VQA~\citep{tu2023unkvqa}. 

\noindent\textbf{Evaluation on TUBench.}
TUBench presents a formidable out-of-distribution challenge because its questions and images are collected using distinct protocols across multiple specialized subdomains. Table~\ref{tab:unified_ood_combined} shows that Rep2Act achieves superior performance among all evaluated 3B-scale methods, with $54.21\%$ answerability-classification accuracy and $56.84\%$ response-level A-Acc. 
We further evaluate Rep2Act using the official subset-specific prompts and parsers provided by TUBench, as shown in Figure~\ref{fig:tubench_official_radar}, with complete results in Appendix~\ref{sec:appendix_tubench_official}. Following~\citet{yu2024tubench}, F1 is computed by treating the unanswerable class as the positive class. Rep2Act-3B substantially outperforms all open-source baselines and remains competitive with frontier proprietary models. 

\noindent\textbf{Evaluation on UNK-VQA.}
As shown in Table~\ref{tab:unified_ood_combined}, Rep2Act achieves superior performance among all evaluated 3B-scale methods with answerability-classification accuracy of $55.37\%$ and A-Acc of $55.38\%$. 
We further evaluate Rep2Act under the official UNK-VQA protocol in Binary (BY) and Multiple-Choice (MC) settings, where the prompt templates provide four candidate answer options for each question. Table~\ref{tab:published_ood_explicit_prompt} shows that Rep2Act performs on par with Instruct-BLIP-Vicuna-13B in BY setting. In MC setting, Rep2Act improves upon 3B base model by $9.54\%$ and outperforms all evaluated baselines despite their considerably larger parameters. These results demonstrate that \sys{} generalizes effectively to out-of-distribution data.

\begin{table*}[t]
\begin{minipage}[t]{0.48\textwidth}
\caption{Zero-shot evaluation on UNK-VQA under judgment and multiple-choice settings.}\vspace{-0.1in}
\label{tab:published_ood_explicit_prompt}
\centering
\setlength{\tabcolsep}{6pt}
\scriptsize
\begin{tabular}{lcc}
\toprule
\textbf{Model} & \textbf{BY Acc} & \textbf{MC Acc} \\
\midrule
Otter-MPT-7B & 40.87 & 20.33 \\
Open-Flamingo-MPT-7B & 31.49 & \underline{26.52} \\
Instruct-BLIP-Vicuna-7B & 43.98 & 22.38 \\
Instruct-BLIP-Vicuna-13B & \textbf{54.82} & 18.41 \\
\midrule
Qwen2.5-VL-3B & 52.44 & 18.68 \\
Rep2Act 3B & \underline{54.78} & \textbf{28.22} \\
\bottomrule
\end{tabular}
\end{minipage}\hfill
\begin{minipage}[t]{0.48\textwidth}
\caption{VQA capabilities preservation on POPE, GQA and TextVQA.}\vspace{-0.1in}
\label{tab:general_vqa_results}
\centering
\setlength{\tabcolsep}{6pt}
\scriptsize
\begin{tabular}{lccc}
\toprule
\textbf{Model} & \textbf{POPE} & \textbf{GQA} & \textbf{TextVQA} \\
\midrule
Qwen2.5-VL-3B  & \textbf{86.04} & \textbf{60.09} & \textbf{79.56} \\
+ Margin Loss & \underline{85.95} & 59.64 & \underline{79.40} \\
+ Route-Contrastive & 85.74 & 59.56 & 79.08 \\
Rep2Act & 85.85 & \underline{59.71} & 79.25 \\
\midrule
Qwen2.5-VL-7B & \textbf{86.60} & \textbf{60.98} & \textbf{83.75} \\
Rep2Act & \underline{86.11} & \underline{60.34} & \underline{83.62} \\
\bottomrule
\end{tabular}
\end{minipage}\vspace{-0.1in}
\end{table*}

\begin{table*}[t]
\begin{minipage}[t]{0.495\textwidth}
\caption{Response-level component ablation on VAD-R. Cons. denotes route-action consistency.}\vspace{-0.1in}
\label{tab:loss_hyperparam_ablation}
\centering
\setlength{\tabcolsep}{2.5pt}
\scriptsize
\resizebox{\linewidth}{!}{%
\begin{tabular}{lccccccc}
\toprule
\textbf{Configuration} & \textbf{$\lambda_m$} & \textbf{$\lambda_c$} & \textbf{A-Acc} & \textbf{M-P} & \textbf{M-R} & \textbf{M-F1} & \textbf{Cons.} \\
\midrule
w/o Margin \& RC & 0.0 & 0.0 & 73.00 & 73.23 & 73.00 & 72.93 & 95.67 \\
+ Margin Loss & 0.2 & 0.0 & \underline{78.33} & \underline{78.55} & \underline{78.33} & \underline{78.29} & 89.00 \\
+ Route-Contrastive & 0.0 & 1.0 & 78.00 & 78.51 & 78.00 & 77.90 & \textbf{99.67} \\
Rep2Act Combined & 0.2 & 1.0 & \textbf{86.33} & \textbf{86.33} & \textbf{86.33} & \textbf{86.33} & \underline{99.33} \\
\bottomrule
\end{tabular}%
}
\end{minipage}%
\hfill
\begin{minipage}[t]{0.47\textwidth}
\caption{Response-level comparison between FA-Route-SFT and Rep2Act on VAD-R.}\vspace{-0.1in}
\label{tab:route_feature_ablation}
\centering
\setlength{\tabcolsep}{3pt}
\scriptsize
\resizebox{\linewidth}{!}{%
\begin{tabular}{lccccc}
\toprule
\textbf{Method} & \textbf{A-Acc} & \textbf{M-P} & \textbf{M-R} & \textbf{M-F1} & \textbf{C-Acc} \\
\midrule
Route-SFT & 76.67 & \underline{78.50} & 76.67 & 76.29 & 49.67 \\
FA-Route-SFT & \underline{77.33} & 78.47 & \underline{77.33} & \underline{77.10} & \underline{52.67} \\
Rep2Act & \textbf{86.33} & \textbf{86.33} & \textbf{86.33} & \textbf{86.33} & \textbf{57.00} \\
\bottomrule
\end{tabular}%
}
\end{minipage}\vspace{-0.1in}
\end{table*}

\subsection{Ablation Study (RQ3)}
\label{sec:rq4_ablations}

\noindent\textbf{Effectiveness of Loss Design.}
We vary $\lambda_m$ and $\lambda_c$ in Equation~\ref{eq:total_loss}, while retaining the language-modeling and representation-based routing-head supervision in every trained variant and fixing the inference-time fusion weight at $\alpha=0.50$. As shown in Table~\ref{tab:loss_hyperparam_ablation}, the signed margin loss improves A-Acc by $5.33\%$ over Rep2Act w/o both loss. The route-contrastive loss increases route–action consistency to $99.67\%$ and improves A-Acc by $5\%$ . Combining both objectives yields A-Acc of $86.33\%$, representing a $13.33\%$ improvement over Rep2Act w/o either loss. Complete sensitivity analyses for $\lambda_m$ and $\lambda_c$ are provided in Appendix~\ref{sec:appendix_loss_sensitivity}. 

% scores the same gold continuation under the gold and inverted routes,

\noindent\textbf{Discrete Routing vs.\ Continuous Feature Anchoring.}
We evaluate Feature-Anchor Route-SFT (FA-Route-SFT), which combines $\mathcal{L}_{\mathrm{LM}}$ with $\mathcal{L}_{\mathrm{anchor}}$ (Appendix~\ref{sec:appendix_centroid_geometry}). As shown in Table~\ref{tab:route_feature_ablation}, feature anchoring improves A-Acc by only $0.66\%$ over Route-SFT. Because prompt switching induces a substantial representation shift orthogonal to the answerability direction, optimizing internal projections influences the final A/U token boundary only indirectly. In contrast, \sys{} directly optimizes this decision boundary and trains subsequent generation to remain consistent with the selected route. Appendix~\ref{sec:appendix_centroid_geometry} provides the complete geometric decomposition and gradient derivations.
We additionally evaluate alternative SFT response formats on VAD-R, including concise and detailed targets both w/ and w/o chain-of-thought reasoning. The results are provided in Appendix~\ref{sec:appendix_response_targets}.

% We evaluate Feature-Anchor Route-SFT (FA-Route-SFT), which combines $\mathcal{L}_{\mathrm{LM}}$ with $\mathcal{L}_{\mathrm{anchor}}$ (Appendix~\ref{sec:appendix_centroid_geometry}).
% Table~\ref{tab:route_feature_ablation} shows that feature anchoring improves A-Acc by only 0.66\% over Route-SFT. Because prompt switching induces a large orthogonal representation shift, optimizing internal projections affects the final A/U token boundary indirectly, whereas Rep2Act directly optimizes this boundary and trains downstream generation to adhere to the selected route. Appendix~\ref{sec:appendix_centroid_geometry} provides the complete geometric decomposition and gradient derivations.
% We also evaluate different SFT response formats on VAD-R, using concise or detailed targets with and without chain-of-thought reasoning; see Appendix~\ref{sec:appendix_response_targets}.

\section{Conclusion}
\label{sec:conclusion}

In this work, we investigate the abstention capabilities of VLMs. We construct VAD-R, a genuinely challenging benchmark, and reveal that deep hidden representations clearly distinguish answerable from unanswerable queries, yet this internal signal fails to guide direct generation, exposing a representation-to-action disconnect. To bridge this gap, we introduce \sys{}, which converts latent answerability into an explicit routing decision and aligns subsequent generation with the selected answer-or-abstain action. Extensive evaluations demonstrate that Rep2Act substantially improves both action and content accuracy, generalizes effectively to out-of-distribution benchmarks, and preserves standard VQA capabilities without inducing excessive refusal.

\newpage

\bibliography{iclr2027_conference}
\bibliographystyle{iclr2027_conference}

\newpage
\appendix

\section{Related Work}
\label{sec:related_work}

\paragraph{VLM Abstention Benchmarks.}
% Existing benchmarks study visual unanswerability through naturally difficult images or synthetic evidence removal. 
Several benchmarks have been developed to evaluate the abstention capabilities of VLMs. VizWiz~\citep{gurari2018vizwiz} collects photographs captured by visually impaired users, which often exhibit authentic image-acquisition challenges. 
In contrast, UNK-VQA, MM-UPD and TUBench construct unanswerable inputs through synthetic interventions such as masking, blanking, answer-option deletion, or cross-source mismatching~\citep{tu2023unkvqa,yu2024tubench,zheng2024mmupd}. 
Heuristic dataset construction can introduce visual or linguistic shortcuts that allow answerability labels to be inferred without genuine image–question reasoning~\citep{goyal2017making,agrawal2018dont,geirhos2020shortcut}. 
Moreover, these benchmarks provide only binary answerability labels and lack step-by-step explanations identifying the missing evidence in unanswerable examples. 
Their evaluation protocols explicitly ask models to identify unanswerable inputs and therefore measure prompted judgment rather than spontaneous abstention during direct answering. 
In this work, we introduce VAD-R to address the limitations of existing benchmarks. Each VAD-R sample includes an answerability label and a step-by-step rationale, while unanswerable examples are additionally annotated with causal evidence-gap labels. We employ a two-stage pipeline combining shortcut filtering and quality verification to prevent answerability-label leakage. Furthermore, our evaluation protocol requires models to respond directly without receiving an explicit unanswerability hint, enabling a faithful assessment of spontaneous abstention. 

\paragraph{Methods for Enhancing VLM Abstention Capabilities.}
Recent studies have investigated how to identify unanswerable questions and improve the abstention capabilities of VLMs. Some methods estimate unreliability from response variability, including consistency-based detection~\citep{khan2024consistency}, semantic entropy~\citep{kuhn2023semantic,kuhn2023semantic_nature}, and multimodal perturbation consistency~\citep{zhang2024vluncertainty}. Other approaches exploit internal representations or decoding-time signals: LVLM-LP probes first-token logits~\citep{zhao2024lvlmlp}, CLIP-UP conditions generation on a learned visual prefix~\citep{vardi2026clipup}, and VCD contrasts predictions obtained from clean and corrupted visual inputs~\citep{leng2024vcd}. Although these methods estimate answerability or steer answer-versus-abstain behavior, they do not train models to generate complete responses that explain which visual evidence is missing, occluded, or insufficiently resolved and \sys{} aims to fill this gap.

\clearpage
\section{Supplementary Design and Evaluation Details}
\label{sec:appendix_experimental_details}

\subsection{Detailed Analysis of Unimodal Shortcuts}
\label{sec:appendix_shortcut_probe}

\paragraph{Diagnostic Findings and Comparative Analysis.}
As detailed in Table~\ref{tab:shortcut_probe_full}:
\begin{itemize}[leftmargin=*]
    \item \textbf{Text Shortcuts in Existing Benchmarks}: In MM-UPD, MMAAD achieves an AUROC of $100.00\%$ and MMIASD achieves $99.77\%$ purely from question texts, as negative samples were constructed by removing options or using standardized textual templates. In TUBench, UGeoQA and UVQA reach text-only AUROCs of $84.29\%$ and $84.91\%$. Models can thus identify unanswerability from superficial phrasing without verifying visual evidence.
    \item \textbf{Image Shortcuts in Existing Benchmarks}: In MM-UPD's MMIVQD subset, randomly pairing images from different source distributions produces an Image AUROC of $83.99\%$. In TUBench's UTabMWP, visual masking artifacts yield a perfect Image AUROC of  $100.00\%$.
    \item \textbf{VAD-R Shortcut Diagnosis}:  On the final VAD-R test split, question-only accuracy is $48.22\%$ with AUROC $48.05\%$, and image-only accuracy is $47.56\%$ with AUROC $47.15\%$. These results indicate weak label predictability under the evaluated unimodal probes.
\end{itemize}

\begin{table*}[h]
\caption{\small Unimodal shortcut probing under frozen Qwen2.5-VL-7B representations across UNK-VQA, TUBench, MM-UPD, and VAD-R. Probes are retrained on all 300 questions of VAD-R using five-fold image-grouped cross-validation and three independent runs. Red cells indicate accuracy above $60\%$.}
\label{tab:shortcut_probe_full}
\begin{center}
\resizebox{0.92\textwidth}{!}{%
\begin{tabular}{llcccccc}
\toprule
\textbf{Benchmark / Sub-dataset} & \textbf{Input Modality} & \textbf{Samples} & \textbf{U / A Ratio} & \textbf{Accuracy} & \textbf{Bal-Acc} & \textbf{Macro-F1} & \textbf{AUROC} \\
\midrule
\multicolumn{8}{l}{\textbf{1. UNK-VQA Benchmark}} \\
\textit{Question-Only Probing} & & & & & & & \\
UNK-VQA, Overall & Question-Only & 11,036 & 5,991 / 5,045 & 54.04 & 50.02 & 37.67 & 50.93 $\pm$ 0.07 \\
\quad I-1, Image Context Mislead & Question-Only & 3,522 & 1,998 / 1,524 & 56.73 & 50.46 & 39.93 & 51.78 $\pm$ 0.40 \\
\quad I-2, Object Masking & Question-Only & 807 & 496 / 311 & \cellcolor{red!15}61.46 & 49.42 & 39.68 & 49.30 $\pm$ 1.31 \\
\quad I-3, Object Copy-Move & Question-Only & 812 & 449 / 363 & 55.30 & 49.27 & 37.21 & 47.12 $\pm$ 2.24 \\
\quad T-1, Text Over-Specification & Question-Only & 4,936 & 2,519 / 2,417 & 51.03 & 50.10 & 35.90 & 51.85 $\pm$ 0.42 \\
\quad T-2, Text Missing Details & Question-Only & 959 & 529 / 430 & 55.16 & 49.53 & 37.02 & 49.41 $\pm$ 0.83 \\
\textit{Image-Only Probing} & & & & & & & \\
UNK-VQA, Overall & Image-Only & 2,984 & 1,592 / 1,392 & 52.59 & 51.27 & 49.87 & 53.15 $\pm$ 0.09 \\
\quad I-1, Image Context Mislead & Image-Only & 946 & 530 / 416 & 56.03 & 49.81 & 48.03 & 49.82 $\pm$ 0.35 \\
\quad I-2, Object Masking & Image-Only & 207 & 138 / 69 & 51.09 & 46.45 & 58.92 & 64.89 $\pm$ 1.31 \\
\quad I-3, Object Copy-Move & Image-Only & 208 & 109 / 99 & 50.91 & 45.05 & 60.12 & 56.68 $\pm$ 2.24 \\
\quad T-1, Text Over-Specification & Image-Only & 1,351 & 657 / 694 & 51.37 & 50.61 & 49.67 & 51.39 $\pm$ 0.31 \\
\quad T-2, Text Missing Details & Image-Only & 272 & 158 / 114 & 58.09 & 51.66 & 51.25 & 51.30 $\pm$ 0.45 \\
\midrule
\multicolumn{8}{l}{\textbf{2. TUBench Benchmark}} \\
\textit{Question-Only Probing} & & & & & & & \\
TUBench, Overall & Question-Only & 2,354 & 1,151 / 1,203 & \cellcolor{red!15}68.71 & 68.77 & 68.70 & 77.23 $\pm$ 0.04 \\
\quad UCR, Code Rendering & Question-Only & 480 & 214 / 266 & 57.29 & 55.44 & 54.56 & 58.67 $\pm$ 0.15 \\
\quad UGeoQA, Geometry Reasoning & Question-Only & 974 & 487 / 487 & \cellcolor{red!15}77.24 & 77.24 & 77.23 & 84.29 $\pm$ 0.08 \\
\quad UTabMWP, Table Math & Question-Only & 400 & 200 / 200 & 50.00 & 50.00 & 44.12 & 50.00 $\pm$ 0.00 \\
\quad UVQA, Visual QA & Question-Only & 500 & 250 / 250 & \cellcolor{red!15}77.93 & 77.93 & 77.93 & 84.91 $\pm$ 0.11 \\
\textit{Image-Only Probing} & & & & & & & \\
TUBench, Overall & Image-Only & 2,354 & 1,151 / 1,203 & \cellcolor{red!15}60.88 & 60.77 & 60.65 & 69.11 $\pm$ 0.12 \\
\quad UCR, Code Rendering & Image-Only & 480 & 214 / 266 & \cellcolor{red!15}61.88 & 59.66 & 58.96 & 64.32 $\pm$ 0.21 \\
\quad UGeoQA, Geometry Reasoning & Image-Only & 974 & 487 / 487 & 50.00 & 50.00 & 49.79 & 50.00 $\pm$ 0.00 \\
\quad UTabMWP, Table Math & Image-Only & 400 & 200 / 200 & \cellcolor{red!15}99.67 & 99.67 & 99.67 & 100.00 $\pm$ 0.00 \\
\quad UVQA, Visual QA & Image-Only & 500 & 250 / 250 & 50.13 & 50.13 & 47.45 & 50.02 $\pm$ 0.06 \\
\midrule
\multicolumn{8}{l}{\textbf{3. MM-UPD Benchmark}} \\
\textit{Question-Only Probing} & & & & & & & \\
MM-UPD, ALL Combined & Question-Only & 15,245 & 7,213 / 8,032 & \cellcolor{red!15}90.17 & 89.66 & 89.96 & 94.79 $\pm$ 0.03 \\
\quad MMAAD, Missing Option & Question-Only & 5,537 & 2,359 / 3,178 & \cellcolor{red!15}99.84 & 99.85 & 99.84 & 100.00 $\pm$ 0.00 \\
\quad MMIASD, Inconsistent Annotation & Question-Only & 7,052 & 3,526 / 3,526 & \cellcolor{red!15}98.10 & 98.10 & 98.10 & 99.77 $\pm$ 0.01 \\
\quad MMIVQD, Incompatible Question & Question-Only & 2,656 & 1,328 / 1,328 & 50.01 & 50.01 & 47.36 & 50.01 $\pm$ 0.01 \\
\textit{Image-Only Probing} & & & & & & & \\
MM-UPD, ALL Combined & Image-Only & 15,245 & 7,213 / 8,032 & 54.37 & 53.11 & 50.97 & 55.11 $\pm$ 0.11 \\
\quad MMAAD, Missing Option & Image-Only & 5,537 & 2,359 / 3,178 & 57.40 & 50.00 & 36.47 & 49.81 $\pm$ 0.11 \\
\quad MMIASD, Inconsistent Annotation & Image-Only & 7,052 & 3,526 / 3,526 & 50.00 & 50.00 & 48.79 & 50.00 $\pm$ 0.00 \\
\quad MMIVQD, Incompatible Question & Image-Only & 2,656 & 1,328 / 1,328 & \cellcolor{red!15}74.71 & 74.71 & 74.52 & 83.99 $\pm$ 0.57 \\
\midrule
\multicolumn{8}{l}{\textbf{4. Ours: VAD-R Benchmark}} \\
\textit{Question-Only Probing} & & & & & & & \\
VAD-R & Question-Only & 300 & 150 / 150 & 48.22 & 48.22 & 47.02 & 48.05 $\pm$ 0.68 \\
\quad Counting & Question-Only & 126 & 63 / 63 & 48.68 & 48.68 & 47.59 & 48.37 $\pm$ 1.40 \\
\quad Text \& Symbol & Question-Only & 42 & 21 / 21 & 50.00 & 50.00 & 49.16 & 46.18 $\pm$ 1.82 \\
\quad Visual Attribute & Question-Only & 84 & 42 / 42 & 46.43 & 46.43 & 45.04 & 45.31 $\pm$ 0.60 \\
\quad Visual Localization & Question-Only & 48 & 24 / 24 & 48.61 & 48.61 & 45.23 & 53.07 $\pm$ 0.78 \\
\textit{Image-Only Probing} & & & & & & & \\
VAD-R & Image-Only & 300 & 150 / 150 & 47.56 & 47.56 & 47.51 & 47.15 $\pm$ 0.25 \\
\quad Counting & Image-Only & 126 & 63 / 63 & 48.15 & 48.15 & 48.07 & 45.54 $\pm$ 0.98 \\
\quad Text \& Symbol & Image-Only & 42 & 21 / 21 & 43.65 & 43.65 & 43.17 & 43.12 $\pm$ 2.33 \\
\quad Visual Attribute & Image-Only & 84 & 42 / 42 & 48.81 & 48.81 & 48.72 & 49.39 $\pm$ 0.83 \\
\quad Visual Localization & Image-Only & 48 & 24 / 24 & 47.22 & 47.22 & 47.03 & 50.35 $\pm$ 2.66 \\
\bottomrule
\end{tabular}%
}
\end{center}
\end{table*}

\subsection{Layer-wise Answerability Diagnostics}
\label{sec:appendix_layer_selection}

\paragraph{Answerability Directions and Projection Scores.}
\label{sec:appendix_answerability_definition}
We first examine whether hidden-state representations can distinguish between answerable and unanswerable questions. 
Specifically, we construct two types of prompting modes:
\begin{itemize}[leftmargin=*]
    \item \emph{Direct-Answer Prompt Mode ($P_D$)}, which consists solely of the original question from VAD-R;
    \item \emph{Judgment Prompt Mode ($P_J$)}, which prepends the instruction ``\textit{Based only on the image, can a viewer reliably answer the question? Answer only yes or no.}'' to the original question.
\end{itemize}

Inspired by prior work~\citep{alain2016understanding,burns2022discover,azaria2023internal,liu2023cognitive}, we adopt linear probing to examine the information encoded in VLM representations. 
Formally, a VQA instance is denoted by $x=(\vi,\vq)\in\mathcal{I}\times\mathcal{Q}$, where $\mathcal{I}$ denotes the image space and $\mathcal{Q}$ denotes the space of tokenized questions over vocabulary $\mathcal{V}$. The image $\vi\in\mathbb{R}^{H\times W\times C}$ has height $H$, width $W$, and $C$ channels, while the question $\vq=(w_1^q,\ldots,w_{|\vq|}^q)$ comprises $|\vq|$ tokens, with each $w_j^q\in\mathcal{V}$. 
Let $y(x)\in\mathcal{Y}=\{\mathrm{A},\mathrm{U}\}$ denote the ground-truth answerability label, where $\mathrm{A}$ and $\mathrm{U}$ correspond to answerable and unanswerable, respectively. 
On training data $\mathcal{D}_{\text{train}}$, let $\mathcal{S}_A = \{x \in \mathcal{D}_{\text{train}} \mid y(x) = \text{A}\}$ and $\mathcal{S}_U = \{x \in \mathcal{D}_{\text{train}} \mid y(x) = \text{U}\}$. 

For each prompting mode $M\in\{D,J\}$ and layer $l\in\{0,\ldots,L\}$, where layer $0$ corresponds to the embedding output and layer $L$ to the final Transformer layer, we extract the hidden state of the final prompt token immediately before generation, denoted by $\vh_l^M(x)\in\mathbb{R}^d$, with $d$ representing the hidden-state dimensionality. 
We then average the representations within each class to obtain the corresponding layerwise centroid: 
\begin{equation}
\label{eq:answerability_class_means}
\vmu_{A,l}^M = \frac{1}{|\mathcal{S}_A|} \sum_{x \in \mathcal{S}_A} \vh_l^{M}(x), \qquad \vmu_{U,l}^M = \frac{1}{|\mathcal{S}_U|} \sum_{x \in \mathcal{S}_U} \vh_l^{M}(x)
\end{equation}
For each prompting mode $M\in\{D,J\}$ at layer $l$, we then compute the following terms: 
\begin{itemize}[leftmargin=*]
    \item \emph{Unit Answerability Direction} is defined as the normalized difference between the answerable and unanswerable mode centroids:$\vd_{M,l} = \vmu_{A,l}^M - \vmu_{U,l}^M, \hat{\vd}_{M,l} = \frac{\vd_{M,l}}{\|\vd_{M,l}\|_2}$
    \item \emph{Mode-Midpoint Centroid}: $\vc_{M,l} = \frac{1}{2}\left(\vmu_{A,l}^M + \vmu_{U,l}^M\right)$
\end{itemize}

Then, we conduct the following analysis:

$\bullet$~\textit{Intra-Mode Analysis.} We derive standardized projection score as follows:
\begin{equation}
\label{eq:answerability_projection_scale}
z_{M,l}(x) = \frac{(\vh_l^{M}(x) - \vc_{M,l})^\top \hat{\vd}_{M,l}}{\sigma_{M,l}}, \quad \sigma_{M,l} = \sqrt{\frac{1}{|\mathcal{D}_{\text{train}}|-1} \sum_{x \in \mathcal{D}_{\text{train}}} \left((\vh_l^{M}(x) - \vc_{M,l})^\top \hat{\vd}_{M,l}\right)^2}
\end{equation}
The standardized score $z_{M,l}(x)$ quantifies the position of a sample’s hidden-state representation along the answerability direction, centered at the midpoint between the two mode centroids and normalized by the sample standard deviation $\sigma_{M,l}$ of the centered training projections.
Positive values indicate alignment with the answerable class, whereas negative values indicate alignment with the unanswerable class, and larger absolute values reflect greater separation from the decision boundary. 

$\bullet$~\textit{Inter-Mode Analysis.} We then evaluate whether the answerability direction estimated under the judgment prompting mode can also separate the intermediate representations elicited under the direct-answer mode. Successful transfer would indicate that both prompting modes share a common representation of answerability that can be accessed without an explicit judgment instruction. 

Formally, we derive the following cross-mode score: 
\begin{equation}
\label{eq:answerability_cross_projection}
z_{D\to J,l}(x)=\frac{(\vh_l^D(x)-\vc_{D,l})^\top\hat{\vd}_{J,l}}{{\sigma_{D\to J,l}}},\,
{\sigma_{D\to J,l}=\sqrt{\frac{1}{|\mathcal D_{\mathrm{train}}|-1}\sum_{x\in\mathcal D_{\mathrm{train}}}\left[(\vh_l^D(x)-\vc_{D,l})^\top\hat{\vd}_{J,l}\right]^2}}
\end{equation}

\paragraph{Development Diagnostics.}

We analyze the last prompt token's hidden state at each layer of the frozen Qwen2.5-VL-3B-Instruct and Qwen2.5-VL-7B-Instruct models using 1,226 training and 326 development questions. Projection directions, centroids, and scales are estimated on the training split. We also standardize features using training statistics and fit a class-balanced logistic regression probe per layer and prompt. We select its inverse regularization strength $C_{\mathrm{probe}}\in\{0.01,0.1,1,10\}$ by development Macro-F1 and plot the resulting development scores.

\begin{figure}[h]
\centering
\includegraphics[width=\textwidth]{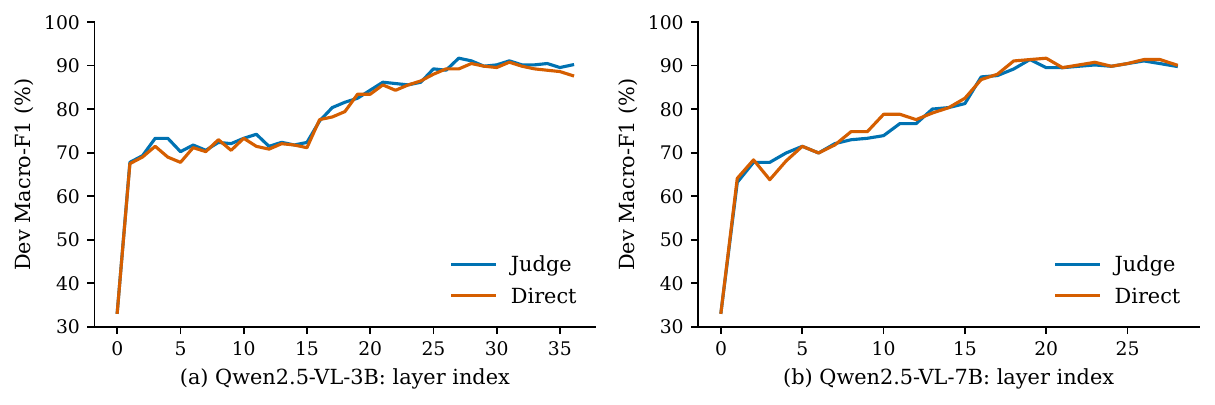}
\caption{\small Layerwise linear-probe macro-F1 on the development set for (a) Qwen2.5-VL-3B-Instruct and (b) Qwen2.5-VL-7B-Instruct. The probes operate on frozen hidden states extracted from the final prompt token, with probe parameters fitted on the training set and regularization hyperparameters selected on the development set. The horizontal axes indicate hidden-state indices, where index 0 denotes the embedding output. Blue and orange represent the judgment and direct-answer prompting modes, respectively. Higher values indicate stronger discrimination between answerable and unanswerable questions..}
\label{fig:layer_selection_diagnostics}
\end{figure}

Figure~\ref{fig:layer_selection_diagnostics} reveals stronger answerability separation in the later layers under both prompting modes. For the 3B model, the judgment-mode and direct-answer probes achieve peak development-set macro-F1 scores of $91.72\%$ at layer 27 and $90.79\%$ at layer 31, respectively. For the 7B model, the corresponding maxima are $91.40\%$ at layer 19 and $91.71\%$ at layer 20. These development-set diagnostics support the use of later-layer representations in the routing head. 

\begin{figure*}[h]
\centering
\includegraphics[width=\textwidth]{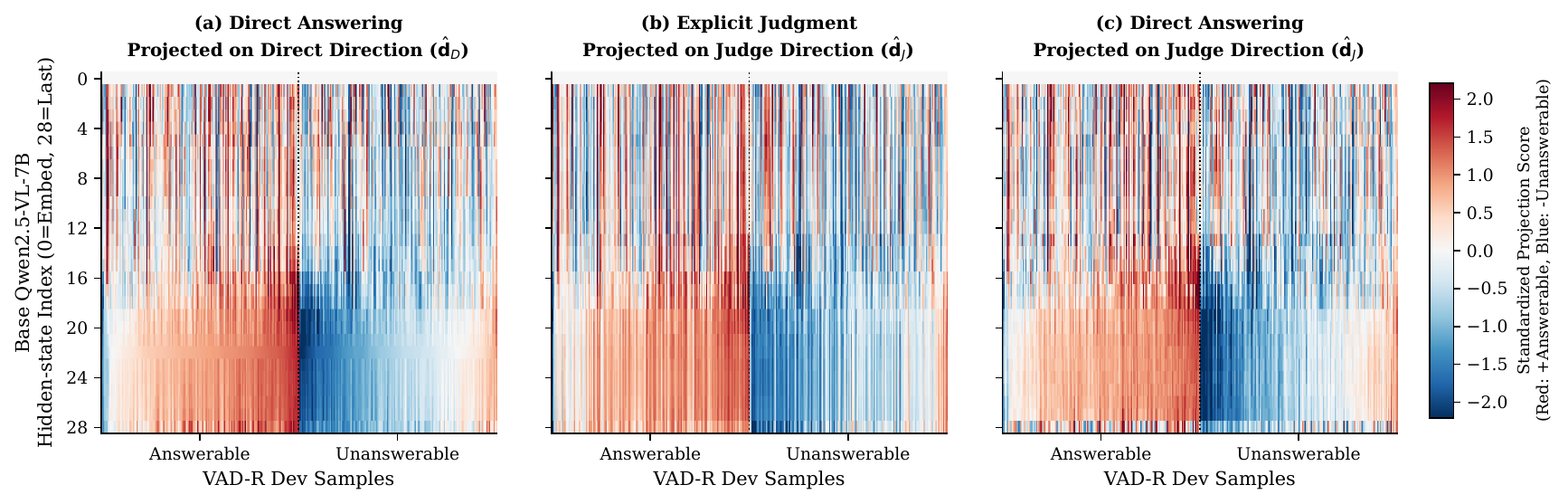}
\caption{\small Layerwise per-sample answerability-projection heatmaps for Qwen2.5-VL-7B on the VAD-R development split, comprising 163 answerable instances on the left and 163 unanswerable instances on the right. The answerability directions, class centroids, and normalization scales are estimated from the training split. \textbf{(a)} Direct-answer representations projected onto their intrinsic answerability direction $\hat{\vd}_{D,l}$. \textbf{(b)} Explicit-judgment representations projected onto the judgment-mode answerability direction $\hat{\vd}_{J,l}$. \textbf{(c)} Direct-answer representations projected onto the judgment-mode direction $\hat{\vd}_{J,l}$ after centering by $\vc_{D,l}$ and normalizing by $\sigma_{D\to J,l}$. Positive values, shown in red, indicate alignment with the answerable direction, whereas negative values, shown in blue, indicate alignment with the unanswerable direction.}
\label{fig:base7b_heatmap}
\end{figure*}

Figure~\ref{fig:base7b_heatmap} presents the 7B projection heatmaps for the same 326 development examples used in Figure~\ref{fig:base3b_heatmap}. At layer 22, which achieves the highest development-set AUROC for the direct-answer projection, the direct-answer, judgment-mode, and cross-prompt projections attain AUROCs of $94.72\%$, $94.51\%$, and $93.98\%$, respectively. These results demonstrate that the cross-prompt projection retains strong answerability separation at the 7B scale. 

\subsection{VAD-R Benchmark Construction, Taxonomies, and Distributions}
\label{sec:appendix_vadr_construction}

\paragraph{Benchmark Motivation and Design Principles.}
Vision-language models frequently hallucinate answers to unanswerable VQAs due to strong language priors and training objectives that reward answering regardless of visual evidence sufficiency. Existing unanswerability datasets predominantly rely on coarse binary labels or fixed refusal phrases, lacking fine-grained evidence grounding or verifiable causal explanations. We design VAD-R as a diagnostic evaluation benchmark. Every question is annotated with a step-by-step visual reasoning rationale, and every unanswerable question includes an explicit explanation detailing precisely why the visual evidence is insufficient, mapped to a causal taxonomy.

\paragraph{Three-Stage Construction Pipeline.}
 We first generate candidate questions, then apply shortcut filtering and independent quality verification:
\begin{enumerate}[leftmargin=*]
    \item \textbf{Stage 1: Qwen3.5-122B-A10B Generation.} Over a diverse collection of real-world images from DOCCI~\citep{onoe2024docci} spanning four core visual domains, namely Fine-Grained Counting, Text \& Symbol Reading, Visual Attribute Recognition, and Visual Localization, Qwen3.5-122B-A10B~\citep{qwen3.5} generates candidate questions across both answerable and unanswerable queries referencing the visual scenes.  Every candidate question receives a step-by-step reasoning rationale, and unanswerable candidates additionally receive causal evidence-gap labels. Initial quality filtering yields 30,386 questions.
    \item \textbf{Stage 2: Question-Only Difficulty Filtering with Qwen2.5-VL-7B-Instruct (Top 1/3).} A common vulnerability in VLM unanswerability benchmarks is language shortcut leakage, where models predict unanswerability purely from text phrasing without inspecting visual evidence. To eliminate textual shortcuts, we train a two-layer Question-Only MLP probe on frozen Qwen2.5-VL-7B-Instruct hidden features extracted at the final prompt token. The difficulty of each question is quantified by its cross-entropy loss under the probe, where a higher loss signifies that the answerability label cannot be easily predicted from text alone.  Specifically, the Question-Only MLP probe is trained exclusively on the candidate pool of the training split ($\mathcal{D}_{\mathrm{train}}$). To establish an unbiased filtering criterion, we evaluate probe cross-entropy losses on the development split ($\mathcal{D}_{\mathrm{dev}}$) to calibrate category-specific score thresholds that retain approximately the hardest top one-third ($1/3$) of candidate questions within each visual dimension and answerability category. Finally, we apply these fixed thresholds uniformly across candidate questions in the training, development, and test splits ($\mathcal{D}_{\mathrm{train}}$, $\mathcal{D}_{\mathrm{dev}}$, and $\mathcal{D}_{\mathrm{test}}$). This procedure retains 8,357 questions in total, including 4,066 unanswerable candidates, purging textual shortcuts while preventing distribution leakage across splits.
    \item \textbf{Stage 3: Qwen3.6-27B Quality Verification and Class Balancing.} To ensure that unanswerable queries genuinely lack sufficient visual evidence, Qwen3.6-27B~\citep{qwen3.6} independently evaluates each unanswerable candidate using the image, question, and candidate reference explanation, assigning a continuous verification confidence score in $[0, 1]$. We strictly retain candidates with verification scores of at least 0.8. After selecting the unanswerable questions, we retain an equal number of answerable questions, enforcing a balanced 1:1 ratio between answerable and unanswerable questions. Among the 4,066 audited unanswerable candidates, 926 pass verification, and pairing them with an equal number of answerable questions yields a balanced benchmark of 1,852 questions. 
\end{enumerate}
This pipeline produces the final benchmark of 1,852 questions, partitioned into 1,226 training, 326 development, and 300 testing questions.

\paragraph{Human Quality Audit.}
To validate annotation quality, an independent human expert conducted a stratified audit on 100 randomly sampled test questions across all four visual skills and answerability classes. The annotator independently verified whether the image genuinely supports the answer or substantiates the stated reason for unanswerability. Across answerable and unanswerable questions combined, 93 annotations were confirmed completely correct, 6 were incorrect, and 1 was borderline, yielding a confirmed accuracy of 93\%.

\paragraph{Visual Dimensions and Causal Taxonomies.}
Table~\ref{tab:vadr_distribution} provides the detailed quantitative distributions of VAD-R across visual dimensions and causal evidence-gap categories.
Subtable (a) details the distribution across 4 visual dimensions: Fine-Grained Object Counting (814 questions, 43.95\%), Visual Attribute Recognition (512 questions, 27.65\%), Visual Localization (268 questions, 14.47\%), and Text \& Symbol Reading (258 questions, 13.93\%).
Subtable (b) presents the distribution across 6 causal evidence-gap categories for the 926 unanswerable questions: \textit{Too Small or Low Resolution} (52.38\%), \textit{Occluded or Not Visible} (44.28\%), \textit{Cropped or Out of Frame} (11.45\%), \textit{Unreadable or Indistinct Text} (11.12\%), \textit{Not Present} (9.50\%), and \textit{Requires External or Nonvisual Information} (1.73\%). Multiple causal categories may apply to a single question.
The balanced 300-question test split contains 150 answerable and 150 unanswerable queries.

\begin{table}[t]
\caption{Distributional breakdown of the VAD-R benchmark across visual dimensions and causal evidence-gap categories. Percentages in Subtable (b) are calculated relative to the 926 unanswerable questions. Multiple labels may apply to a single question.}
\label{tab:vadr_distribution}
\centering
\begin{subtable}[t]{\linewidth}
\caption{Distribution across visual skills ($N=1{,}852$).}
\label{tab:vadr_skills_distribution}
\centering
\small
\resizebox{\linewidth}{!}{%
\begin{tabular}{lccl}
\toprule
\textbf{Visual Dimension} & \textbf{Count} & \textbf{Ratio (\%)} & \textbf{Scope / Description} \\
\midrule
Fine-Grained Counting & 814 & 43.95 & Number of objects, including small, overlapping, or densely arranged instances \\
Visual Attribute & 512 & 27.65 & Queries about color, texture, material, shape, and fine-grained visual states \\
Visual Localization & 268 & 14.47 & Relative positions, spatial orientation, and coordinates of objects within a scene \\
Text \& Symbol Reading & 258 & 13.93 & Scene text, logos, icons, alphanumeric markings, and typography \\
\midrule
\textbf{Total} & \textbf{1,852} & \textbf{100.00} & Complete set of benchmark questions across training, dev, and test splits \\
\bottomrule
\end{tabular}%
}
\end{subtable}

\vspace{0.8em}

\begin{subtable}[t]{\linewidth}
\caption{Distribution across causal evidence-gap categories for unanswerable queries ($N_{\mathrm{U}}=926$).}
\label{tab:vadr_evidence_gap_distribution}
\centering
\small
\resizebox{\linewidth}{!}{%
\begin{tabular}{lccl}
\toprule
\textbf{Evidence-Gap Category} & \textbf{Count} & \textbf{Ratio (\%)} & \textbf{Scope / Description} \\
\midrule
Too Small / Low Resolution & 485 & 52.38 & The relevant target region lacks sufficient resolution or detail for a reliable answer \\
Occluded / Not Visible & 410 & 44.28 & The queried object, feature, or attribute is physically blocked or not visible \\
Cropped / Out of Frame & 106 & 11.45 & The required visual entity or region extends beyond the image boundaries \\
Unreadable / Indistinct Text & 103 & 11.12 & Blurred, corrupted, low-contrast, or incomplete text prevents reliable transcription \\
Not Present & 88 & 9.50 & The queried entity is completely absent from the visual scene \\
External / Nonvisual Info & 16 & 1.73 & The query requires domain facts or external knowledge not depicted in the image \\
\bottomrule
\end{tabular}%
}
\end{subtable}
\end{table}

\paragraph{External Training Datasets.}
Joint instruction tuning uses 1,226 VAD-R training examples and 3,406 external samples, comprising 1,703 answerable examples from ScienceQA~\citep{lu2022scienceqa} and 1,703 unanswerable examples from VizWiz~\citep{gurari2018vizwiz}. Qwen3.6-27B generates the supervision targets for both external sources as follows:

\noindent$\bullet$~\emph{ScienceQA}: we convert image-containing multiple-choice examples into open-ended visual questions with short text answers and reasoning rationales. The conversion model receives the image, original question, choices, correct answer, and available hints, lectures, and solutions. We retain examples verified to be answerable strictly from the visual scene and rewritten query, filtering out instances that depend on external knowledge or omitted options.

\noindent$\bullet$~\emph{VizWiz}: we start from the official training split, requiring at least 7 of 10 annotators to vote unanswerable. Qwen3.6-27B assigns evidence-gap labels, writes an image-grounded rationale, and generates a natural refusal. Invalid annotations are excluded. To balance the external training data, we retain all 1,310 eligible examples with 9--10 votes and select another 393 from the 7--8-vote pool without replacement using a deterministic seed-42 ranking, yielding 1,703 examples.

For external examples, the routing tokens are masked during loss computation, allowing the models to preserve general instruction-following capabilities while learning routed abstention from VAD-R. 

\subsection{Evaluation Benchmarks\texorpdfstring{}{}}
\label{sec:appendix_datasets}

\paragraph{In-Domain Benchmark (VAD-R).}
VAD-R serves as our primary in-domain benchmark for evaluating whether models can spontaneously recognize evidence insufficiency and abstain without prompt hints. As detailed in Appendix~\ref{sec:appendix_vadr_construction}, the evaluation uses the balanced test split of 300 questions (150 answerable and 150 unanswerable), systematically assessing models across answerability classification, A-Acc, and C-Acc.

\paragraph{OOD Generalization Suite.}
To evaluate whether abstention capabilities generalize across diverse domains and synthetic perturbations without retraining, we evaluate on two prominent OOD unanswerability benchmarks:
\begin{itemize}[leftmargin=*]
    \item \textbf{TUBench}~\citep{yu2024tubench}: Evaluates unanswerable question identification across four specialized multimodal reasoning domains: code-based reasoning (UCR, 480 questions), planar geometry problems (UGeoQA, 974 questions), tabular mathematical word problems (UTabMWP, 400 questions), and multimodal visual question answering (UVQA, 500 questions). The complete test split contains 2,354 samples, comprising 1,203 answerable and 1,151 unanswerable questions.
    \item \textbf{UNK-VQA}~\citep{tu2023unkvqa}: Evaluates abstention capabilities across heuristic textual and visual perturbations applied to MS-COCO images and questions,  with five subsets: Image Context Mislead (I-1), Object Masking (I-2), Object Copy-Move (I-3), Text Over-Specification (T-1), and Text Missing Details (T-2).  The evaluated subset contains 4,539 answerable and 5,293 unanswerable questions.

The evaluated UNK-VQA benchmark comprises 9,832 questions with available image assets, supplemented with original MS-COCO images.
\end{itemize}

\paragraph{General Multimodal Perception Benchmarks.}
To verify that our abstention training preserves core visual-understanding capabilities, we evaluate the trained models on three standard vision-language benchmarks:
\begin{itemize}[leftmargin=*]
    \item \textbf{POPE}~\citep{li2023pope}: Probes object hallucination on MS-COCO across three splits of increasing difficulty: Random (sampling non-existent objects uniformly), Popular (sampling frequently occurring co-present objects), and Adversarial (sampling objects that frequently co-occur according to segmentation statistics). It evaluates $9{,}000$ binary yes/no questions, measuring whether models falsely affirm non-existent entities under hallucination pressure.
    \item \textbf{GQA}~\citep{hudson2019gqa}: Evaluates multi-step compositional spatial reasoning, relational scene graph navigation, and structural logic over real-world imagery using $12{,}578$ balanced validation questions, assessing whether abstention mechanisms interfere with complex multi-hop visual reasoning.
    \item \textbf{TextVQA}~\citep{singh2019textvqa}: Assesses optical character recognition (OCR) and open-ended multimodal question answering on $5{,}000$ validation questions across $3{,}166$ images containing scene text from the Open Images dataset, scored via standard soft VQA accuracy across 10 human annotations.
\end{itemize}

\subsection{Baseline Implementation Details}
\label{sec:appendix_baselines}

\begin{itemize}[leftmargin=*]
    \item \textbf{Direct Answering ($P_D$).} Generates a concise response directly from the image and question without explicitly assessing answerability.
    \item \textbf{Judge-then-Answer ($P_{JA}$).} First assesses whether the available visual evidence is sufficient and then either answers the question or explains why a reliable answer cannot be determined, all within a single response. The complete prompts for both baselines are provided in Appendix~\ref{sec:appendix_prompts_baselines}.
    \item \textbf{Semantic Entropy (SE)~\citep{kuhn2023semantic}.} SE samples $K=10$ candidate responses and groups semantically equivalent answers using DeBERTa-v2-xlarge~\citep{deberta}, fine-tuned on MultiNLI for natural language inference (NLI)~\citep{nli}. Two responses are assigned to the same semantic group only if the NLI model predicts entailment in both directions. The semantic entropy computed over these groups is then used for answerability classification. 
    \item \textbf{VL-Uncertainty (VL-U)~\citep{zhang2024vluncertainty}.} Estimates uncertainty from responses sampled under clean inputs, textual paraphrases, and joint text–image perturbations. We evaluate VL-U only on answerability classification.
    \item \textbf{Visual Contrastive Decoding (VCD)~\citep{leng2024vcd}.} For classification, we first generate an answer from the original image. We then measure how much its mean token log-probability drops when noise is added to the image, averaging over five independently sampled noise patterns. A threshold selected on the development set converts this score into an answerability prediction. For response generation, VCD contrasts original-image and noisy-image logits at each decoding step.
    \item \textbf{LVLM-LP~\citep{zhao2024lvlmlp}.} Fits a logistic-regression classifier to the first-token vocabulary logits while keeping the VLM frozen. We evaluate both answerability classification and complete response generation. For inputs predicted to be unanswerable, the fixed prefix \textit{``Sorry, this question is unanswerable, because''} is inserted before the model generates the continuation. 
    \item \textbf{CLIP-UP~\citep{vardi2026clipup}.} Computes the elementwise product of the CLIP image and text embeddings, $\mathbf{u}_{\mathrm{CLIP}}=\mathbf{f}_{\mathrm{img}}\odot\mathbf{f}_{\mathrm{text}}$, and projects the resulting representation into a learned soft prefix that guides answer-or-abstain generation. We evaluate its complete generated responses. 
   \item \textbf{Rep2Act (Ours).} Produces both an answerability classification and a complete response conditioned on the selected A/U routing token. Rep2Act, Rep2Act$^{\dagger}$, and Rep2Act$^{\ddagger}$ employ fused routing ($\alpha=0.50$ for the 3B model and $\alpha=0.15$ for the 7B model), routing-head-only inference ($\alpha=1.0$), and vocabulary-only inference ($\alpha=0.0$), respectively. All three variants are evaluated at both the answerability-classification and response-generation levels. 
\end{itemize}

\subsection{\sys{} Implementation Details}
\label{sec:appendix_training_details}

\paragraph{Model Training and Hyperparameters.}
The training set contains 1,226 VAD-R samples and 3,406 external samples from ScienceQA and VizWiz.
For Rep2Act, the representation route head is first trained for three epochs on VAD-R with the backbone frozen. The 3B and 7B joint stages then use two epochs, an effective batch size of 16, and a learning rate of $10^{-5}$ with AdamW and cosine decay. Joint training updates the route head and LoRA attention adapters; the original backbone weights remain frozen. The thresholds are $\gamma=1.0$ for Signed Margin Loss and $\delta=0.1$ for Route-Contrastive Loss, with weights $\lambda_r=1.0$, $\lambda_m=0.2$, and $\lambda_c=1.0$. The head applies per-layer L2 normalization, mean pooling, LayerNorm, and a two-output linear classifier with softmax. Hidden-state indices are 25--35 for Qwen2.5-VL-3B-Instruct and 20--28 for Qwen2.5-VL-7B-Instruct.

\paragraph{Inference Fusion Parameter $\alpha$ Selection and Freezing.}
Inference uses Equation~\ref{eq:ensemble_prob}. The fusion weight $\alpha$ is selected on the 326-question development set via grid search over $\{0, 0.05, \ldots, 1\}$, yielding $\alpha=0.50$ for Rep2Act-3B and $\alpha=0.15$ for Rep2Act-7B. These parameters are fixed and evaluated across all in-domain and out-of-distribution benchmarks. For the controlled 3B loss ablation on VAD-R, $\alpha=0.50$ is likewise fixed for all variants. General VQA uses vocabulary-only routing with $\alpha=0$; the selected A/U token still conditions the continuation.

\paragraph{Evaluation Metrics and Evaluation Protocols.}
We compare answerability predictions with gold A/U labels and report Accuracy, Macro-Precision, Macro-Recall, Macro-F1, and AUROC, using continuous scores for AUROC. Qwen3.6-27B labels each generated response as answering or abstaining; A-Acc measures agreement with the gold label. For C-Acc on VAD-R, Qwen3.8-27B~\citep{qwen3.8} compares each response with the reference answer and explanation: answerable questions require a correct answer, while unanswerable questions require both abstention and a correct explanation of the missing visual evidence. Official TUBench evaluation uses its subset-specific prompts and parsers and reports F1 for the unanswerable class. For UNK-VQA, we use the published Binary Judgment (BY) and Multiple-Choice (MC) prompts and implement output parsing to compute accuracy. BY predicts answerable or unanswerable; MC selects from three candidate answers and an unanswerable option. Parsing failures count as incorrect. All metrics are reported as percentages.

\paragraph{Hardware and Computing Environment.}
All model fine-tuning, probe training, and generative inference were conducted on a dedicated server equipped with $8\times$ NVIDIA RTX A6000 GPUs, each with 48\,GB of VRAM. The software environment comprised Ubuntu Linux, NVIDIA driver 550.67, CUDA 12.4, PyTorch 2.4, and the vLLM inference engine, which was used to accelerate batched response generation and automated evaluation.

\clearpage
\section{Additional Experimental Results}
\label{sec:appendix_additional_results}

\subsection{Full Evaluation Results Across Individual VLMs on VAD-R}
\label{sec:appendix_full_disconnect_eval}

Table~\ref{tab:eight_model_disconnect_full} presents the complete evaluation results for 8 open-source VLMs and 3 closed-source models on the VAD-R test split. 
Under direct answering ($P_D$), models across parameter scales predominantly attempt to answer every question, achieving near-perfect recall on answerable queries but minimal abstention recall on unanswerable ones. 
Specifically, abstention recall ranges from $0.0\%$ to $23.3\%$ for open-source models and from $11.9\%$ to $22.5\%$ for closed-source models. 
In contrast, when prompted to make an explicit judgment ($P_J$), the same models exhibit substantially higher abstention recall, reaching $96.7\%$ for Qwen3.6-27B and $98.7\%$ for Claude-Haiku-4.5. This striking contrast demonstrates that the representation-to-action disconnect is a pervasive phenomenon across both open-source models and commercial frontier models.

\begin{table*}[h]
\caption{\small Complete empirical evaluation across 8 open-source VLMs and 3 closed-source models on the VAD-R test set under Direct Answering ($P_D$) and Explicit Judgment ($P_J$) modes. Direct responses are classified by Qwen3.6-27B as Action Judge. Metrics include Action Accuracy (A-Acc), Answerable Recall (Ans-R), Abstain Recall on unanswerable queries (Abs-R), Abstain F1 (Abs-F1), and the Abstain Recall Gap $\Delta\text{Abs-R} = \text{Abs-R}_{P_J} - \text{Abs-R}_{P_D}$. A-Acc measures answer/abstain correctness from generated responses under $P_D$ and explicit yes/no judgments under $P_J$. }
\label{tab:eight_model_disconnect_full}
\begin{center}
\resizebox{\textwidth}{!}{%
\begin{tabular}{lccccccccc}
\toprule
& \multicolumn{4}{c}{\textbf{Direct Answering Mode ($P_D$)}} & \multicolumn{4}{c}{\textbf{Explicit Judgment Mode ($P_J$)}} & \textbf{Gap} \\
\cmidrule(lr){2-5}\cmidrule(lr){6-9}\cmidrule(lr){10-10}
\textbf{Model Architecture} & \textbf{A-Acc} & \textbf{Ans-R} & \textbf{Abs-R} & \textbf{Abs-F1} & \textbf{A-Acc} & \textbf{Ans-R} & \textbf{Abs-R} & \textbf{Abs-F1} & \textbf{$\Delta$ Abs-R} \\
\midrule
\multicolumn{10}{l}{\textbf{Open-Source Models}} \\
Qwen2.5-VL-3B & 56.7 & \textbf{100.0} & 13.3 & 23.5 & 74.7 & 81.3 & 68.0 & 73.4 & +54.7 \\
Qwen2.5-VL-7B & 59.3 & \textbf{100.0} & 18.5 & 31.3 & 83.4 & 76.8 & 90.1 & 84.5 & +71.6 \\
Qwen3.6-27B & \textbf{61.7} & \textbf{100.0} & \textbf{23.3} & \textbf{37.8} & \textbf{91.3} & 86.0 & \underline{96.7} & \textbf{91.8} & +73.4 \\
Qwen3.8-27B & \underline{61.0} & \textbf{100.0} & 22.0 & 36.1 & \underline{89.7} & \textbf{93.3} & 86.0 & 89.3 & +64.0 \\
LLaVA-v1.5-7B & 50.3 & \textbf{100.0} & 0.7 & 1.3 & 62.9 & 76.2 & 49.7 & 57.3 & +49.0 \\
LLaVA-v1.5-13B & 50.0 & \textbf{100.0} & 0.0 & 0.0 & 67.2 & 74.2 & 60.3 & 64.8 & +60.3 \\
InternVL3-8B & 54.3 & \textbf{100.0} & 8.6 & 15.9 & 78.1 & 74.8 & 81.5 & 78.9 & +72.9 \\
mPLUG-Owl2-7B & 52.3 & \textbf{100.0} & 4.6 & 8.9 & 73.2 & 55.0 & 91.4 & 77.3 & +86.8 \\
\midrule
\textbf{Open-Source Average} & \textbf{55.7} & \textbf{100.0} & \textbf{11.4} & \textbf{19.4} & \textbf{77.6} & \textbf{77.2} & \textbf{78.0} & \textbf{77.2} & +66.6 \\
\midrule
\multicolumn{10}{l}{\textbf{Closed-Source Models}} \\
Gemini-3.8-Flash & 59.9 & \underline{97.4} & \underline{22.5} & \underline{36.6} & 86.8 & 76.8 & \underline{96.7} & 88.0 & +74.2 \\
Claude-Haiku-4.5 & 55.6 & 96.7 & 14.6 & 24.9 & 68.5 & 38.4 & \textbf{98.7} & 75.8 & +84.1 \\
GPT-5.4-mini & 56.0 & \textbf{100.0} & 11.9 & 21.3 & \underline{89.7} & \underline{90.1} & 89.4 & \underline{89.7} & +77.5 \\
\midrule
\textbf{Closed-Source Average} & \textbf{57.2} & \textbf{98.0} & \textbf{16.3} & \textbf{27.6} & \textbf{81.7} & \textbf{68.4} & \textbf{94.9} & \textbf{84.5} & +78.6 \\
\bottomrule
\end{tabular}%
}
\end{center}
\end{table*}

\paragraph{Fine-Grained Visual Skill Breakdown.}
Table~\ref{tab:skill_breakdown} reports results under both prompting modes across 4 visual skills. For most open-source and closed-source models, explicit judgment yields high abstention recall for text recognition, attribute identification, and localization, whereas direct answering substantially reduces it. Counting is particularly challenging during direct generation: every base model abstains on fewer than $13\%$ of unanswerable counting questions. Localization is comparatively less affected, particularly for Qwen models. These results demonstrate that the representation-to-action disconnect persists across model families and is most pronounced for counting and fine-grained visual details. Rep2Act improves direct-answering accuracy across all skills and achieves the strongest average performance at both model scales. Rep2Act-7B performs best on counting, while Rep2Act-3B leads on the other three skills. 

\begin{table*}[t]
\caption{Fine-grained performance breakdown across four visual skills on VAD-R. A-Acc measures answer/abstain correctness from generated responses under $P_D$ and explicit yes/no judgments under $P_J$.}
\label{tab:skill_breakdown}
\begin{center}
\resizebox{\textwidth}{!}{%
\begin{tabular}{lcccccccccc}
\toprule
& \multicolumn{2}{c}{\textbf{Counting}} & \multicolumn{2}{c}{\textbf{Text \& Symbol}} & \multicolumn{2}{c}{\textbf{Visual Attribute}} & \multicolumn{2}{c}{\textbf{Localization}} & \multicolumn{2}{c}{\textbf{Average}} \\
\cmidrule(lr){2-3}\cmidrule(lr){4-5}\cmidrule(lr){6-7}\cmidrule(lr){8-9}\cmidrule(lr){10-11}
\textbf{Model / Prompting Setup} & \textbf{A-Acc} & \textbf{Abs-R} & \textbf{A-Acc} & \textbf{Abs-R} & \textbf{A-Acc} & \textbf{Abs-R} & \textbf{A-Acc} & \textbf{Abs-R} & \textbf{A-Acc} & \textbf{Abs-R} \\
\midrule
\multicolumn{11}{l}{\textit{Explicit Judgment Mode ($P_J$)}} \\
Qwen2.5-VL-3B & 69.05 & 42.86 & \underline{88.10} & \textbf{100.00} & 78.57 & 69.05 & \textbf{95.83} & \textbf{100.00} & 82.89 & 77.98 \\
Qwen2.5-VL-7B & 82.54 & 77.78 & 85.71 & \textbf{100.00} & 80.95 & \textbf{100.00} & 89.58 & \textbf{100.00} & 84.69 & 94.44 \\
Qwen3.6-27B & 92.06 & 96.83 & \textbf{92.86} & \underline{95.24} & \textbf{91.67} & \underline{97.62} & 89.58 & \textbf{100.00} & \textbf{91.54} & \underline{97.42} \\
Qwen3.8-27B & \underline{93.65} & \underline{98.41} & 85.71 & 76.19 & 83.33 & 69.05 & \textbf{95.83} & \underline{95.83} & 89.63 & 84.87 \\
LLaVA-v1.5-7B & 54.76 & 28.57 & 66.67 & \textbf{100.00} & 61.90 & 45.24 & 83.33 & 70.83 & 66.67 & 61.16 \\
LLaVA-v1.5-13B & 69.05 & 65.08 & 61.90 & 90.48 & 63.10 & 45.24 & 75.00 & 50.00 & 67.26 & 62.70 \\
InternVL3-8B & 74.60 & 68.25 & 80.95 & \textbf{100.00} & 77.38 & 85.71 & 87.50 & \underline{95.83} & 80.11 & 87.45 \\
mPLUG-Owl2-7B & 76.98 & 82.54 & 59.52 & \textbf{100.00} & 63.10 & \textbf{100.00} & \underline{93.75} & \underline{95.83} & 73.34 & 94.59 \\
\cmidrule[0.3pt](lr){1-11}
\textit{Gemini-3.8-Flash} & 84.13 & \underline{98.41} & 85.71 & \underline{95.24} & \underline{89.29} & 92.86 & 89.58 & \textbf{100.00} & 87.18 & 96.63 \\
\textit{Claude-Haiku-4.5} & 65.08 & \textbf{100.00} & 64.29 & \textbf{100.00} & 72.62 & 95.24 & 75.00 & \textbf{100.00} & 69.25 & \textbf{98.81} \\
\textit{GPT-5.4-mini} & \textbf{94.44} & \textbf{100.00} & \textbf{92.86} & \textbf{100.00} & 79.76 & 66.67 & \underline{93.75} & 91.67 & \underline{90.20} & 89.59 \\
\midrule[1.0pt]
\multicolumn{11}{l}{\textit{Direct Answering Mode ($P_D$)}} \\
Qwen2.5-VL-3B & 49.21 & 0.00 & 59.52 & 19.05 & 51.19 & 2.38 & 83.33 & 66.67 & 60.81 & 22.02 \\
Qwen2.5-VL-7B & 55.56 & 11.11 & 54.76 & 9.52 & 53.57 & 7.14 & 83.33 & 66.67 & 61.80 & 23.61 \\
Qwen3.6-27B & 56.35 & 12.70 & 61.90 & 23.81 & 59.52 & 19.05 & 85.42 & 70.83 & 65.80 & 31.60 \\
Qwen3.8-27B & 53.17 & 6.35 & 59.52 & 19.05 & 60.71 & 21.43 & 83.33 & 66.67 & 64.18 & 28.37 \\
LLaVA-v1.5-7B & 50.79 & 1.59 & 50.00 & 0.00 & 50.00 & 0.00 & 50.00 & 0.00 & 50.20 & 0.40 \\
LLaVA-v1.5-13B & 50.00 & 0.00 & 50.00 & 0.00 & 50.00 & 0.00 & 50.00 & 0.00 & 50.00 & 0.00 \\
InternVL3-8B & 50.79 & 1.59 & 54.76 & 9.52 & 51.19 & 2.38 & 68.75 & 37.50 & 56.37 & 12.75 \\
mPLUG-Owl2-7B & 54.76 & 9.52 & 52.38 & 4.76 & 50.00 & 0.00 & 50.00 & 0.00 & 51.79 & 3.57 \\
\cmidrule[0.3pt](lr){1-11}
\textit{Gemini-3.8-Flash} & 53.17 & 7.94 & 73.81 & 47.62 & 60.71 & 26.19 & 64.58 & 33.33 & 63.07 & 28.77 \\
\textit{Claude-Haiku-4.5} & 53.17 & 6.35 & 61.90 & 33.33 & 48.81 & 2.38 & 68.75 & 41.67 & 58.16 & 20.93 \\
\textit{GPT-5.4-mini} & 50.79 & 1.59 & 64.29 & 28.57 & 51.19 & 2.38 & 70.83 & 41.67 & 59.28 & 18.55 \\
\cmidrule[0.3pt](lr){1-11}
\textbf{Rep2Act-3B} & \underline{82.54} & \underline{74.60} & \textbf{85.71} & \textbf{95.24} & \textbf{86.90} & \textbf{90.48} & \textbf{95.83} & \textbf{100.00} & \underline{87.74} & \textbf{90.08} \\
\textbf{Rep2Act-7B} & \textbf{91.27} & \textbf{92.06} & \underline{83.33} & \underline{90.48} & \underline{84.52} & \underline{83.33} & \underline{93.75} & \underline{91.67} & \textbf{88.22} & \underline{89.39} \\
\bottomrule
\end{tabular}%
}
\end{center}
\end{table*}

\subsection{Complete TUBench Results under Official Evaluation Protocol}
\label{sec:appendix_tubench_official}
Table~\ref{tab:tubench_official_subset_comparison} presents the complete numerical results corresponding to Figure~\ref{fig:tubench_official_radar}. Rep2Act-3B achieves an average F1 score of $53.3\%$, outperforming GPT-4 Turbo and GPT-4o by $16.2\%$ and $1.1\%$, respectively, while delivering leading performance on specialized subsets such as UTabMWP. These results demonstrate that explicit representation-to-action routing transfers effectively to diverse out-of-distribution structures without task-specific fine-tuning. 

\begin{table*}[t]
\caption{TUBench OOD evaluation under official subset-specific prompts and parsers. F1 is evaluated on the unanswerable class.}
\label{tab:tubench_official_subset_comparison}
\begin{center}
\setlength{\tabcolsep}{2.2pt}
\scriptsize
\resizebox{\textwidth}{!}{%
\begin{tabular}{lccccccccccccccc}
\toprule
& \multicolumn{3}{c}{\textbf{UCR}} & \multicolumn{3}{c}{\textbf{UVQA}} & \multicolumn{3}{c}{\textbf{UGeoQA}} & \multicolumn{3}{c}{\textbf{UTabMWP}} & \multicolumn{3}{c}{\textbf{Average}} \\
\cmidrule(lr){2-4}\cmidrule(lr){5-7}\cmidrule(lr){8-10}\cmidrule(lr){11-13}\cmidrule(lr){14-16}
\textbf{Model} & \textbf{F1} & \textbf{Acc} & \textbf{C-Acc} & \textbf{F1} & \textbf{Acc} & \textbf{C-Acc} & \textbf{F1} & \textbf{Acc} & \textbf{C-Acc} & \textbf{F1} & \textbf{Acc} & \textbf{C-Acc} & \textbf{F1} & \textbf{Acc} & \textbf{C-Acc} \\
\midrule
\multicolumn{16}{l}{\textbf{Open-Source Models}} \\
LLaVA-v1.5-7B & 0.0 & 55.4 & 28.1 & 0.0 & 50.0 & 31.6 & \underline{63.9} & 51.3 & \textbf{45.0} & 41.8 & 50.5 & 34.2 & 26.4 & 51.8 & 34.7 \\
LLaVA-v1.5-13B & 0.0 & 55.4 & 25.4 & 0.0 & 50.0 & 32.6 & \textbf{64.2} & 51.8 & \textbf{45.0} & 59.2 & 49.2 & 42.2 & 30.8 & 51.6 & 36.3 \\
LLaVA-v1.6-7B & 0.0 & 55.4 & 27.7 & 0.0 & 50.0 & 30.4 & 5.1 & 50.0 & 14.5 & 17.8 & 53.8 & 26.0 & 5.7 & 52.3 & 24.6 \\
LLaVA-v1.6-13B & 16.0 & 54.2 & 29.2 & 7.7 & 51.8 & 30.2 & 48.3 & 48.4 & 30.6 & \underline{61.5} & 53.8 & 45.8 & 33.4 & 52.0 & 33.9 \\
\cmidrule(lr){1-16}
Qwen2.5-VL-3B & 7.0 & 55.4 & 27.9 & 69.1 & 71.4 & 62.0 & 23.2 & 54.5 & 29.1 & 19.0 & 55.3 & 50.5 & 29.6 & 59.2 & 42.4 \\
\midrule
\multicolumn{16}{l}{\textbf{Closed-Source Models}} \\
Qwen-VL-Max & 16.1 & 56.5 & 30.4 & 67.7 & 74.4 & 61.8 & 43.6 & 57.3 & 33.3 & 7.7 & 52.0 & 34.8 & 33.8 & 60.0 & 40.1 \\
Qwen-VL-Plus & 23.8 & 56.0 & 31.5 & 7.6 & 51.6 & 41.0 & 8.8 & 50.8 & 15.4 & 3.9 & 50.7 & 31.8 & 11.0 & 52.3 & 29.9 \\
Qwen3-VL-Flash & 49.2 & \textbf{67.7} & \textbf{57.9} & 48.2 & 65.6 & 58.2 & 9.8 & 52.6 & \textbf{51.2} & \underline{63.5} & \underline{73.2} & \underline{73.0} & 42.7 & 64.8 & \textbf{60.1} \\
Qwen3-VL-Plus & 28.3 & \underline{61.0} & \underline{53.5} & 50.0 & 66.4 & 55.6 & 10.1 & 52.7 & \underline{50.0} & \textbf{72.6} & \textbf{78.5} & \textbf{78.5} & 40.3 & 64.7 & \underline{59.4} \\
GPT-4 Turbo & \underline{57.7} & 57.5 & 45.0 & \underline{77.6} & \textbf{80.6} & \underline{68.4} & 5.6 & 51.4 & 23.2 & 7.7 & 52.0 & 38.8 & 37.1 & 60.4 & 43.8 \\
GPT-4o mini & \textbf{57.8} & 51.9 & 41.0 & \textbf{79.3} & 77.6 & 66.4 & 32.6 & \underline{57.9} & 27.0 & 44.2 & 64.0 & 47.0 & \textbf{53.5} & 62.8 & 45.4 \\
GPT-4o & 53.0 & 60.8 & 39.8 & 76.8 & \underline{80.2} & \textbf{68.6} & 19.5 & 55.1 & 26.6 & 59.6 & 70.5 & 59.5 & 52.2 & \textbf{66.7} & 48.6 \\
\midrule
Rep2Act 3B & 42.1 & 59.4 & 34.4 & 68.6 & 68.2 & 58.0 & 40.1 & \textbf{60.2} & 37.8 & 62.2 & 71.8 & 65.5 & \underline{53.3} & \underline{64.9} & 48.9 \\
\bottomrule
\end{tabular}%
}
\end{center}
\end{table*}

\subsection{SFT Response Target Construction}
\label{sec:appendix_response_targets}

Table~\ref{tab:target_data_ablation} examines the effects of response-target formatting on VAD-R using concise and detailed outputs for both answerable and unanswerable queries. \textit{Short} targets pair concise answers with brief abstention responses such as ``I cannot answer'', whereas \textit{Detailed} targets pair complete answers with explicit causal rationales identifying the missing visual evidence. We evaluate both formats with and without chain-of-thought reasoning. Without an explicit decision token, standard SFT baselines struggle to balance answering and abstaining. \textit{Short w/o CoT} answers nearly every query, achieving $98.00\%$ answer recall but only $28.00\%$ abstention recall. Introducing chain-of-thought reasoning in \textit{Short w/ CoT} increases abstention recall to $73.33\%$ but reduces answer recall to $81.33\%$. Models trained with detailed targets similarly plateau at A-Acc values between $76.67\%$ and $77.67\%$, with abstention recall remaining near $65\%$. In contrast, Rep2Act achieves more balanced performance, with $86.67\%$ answer recall, $86.00\%$ abstention recall, and $86.33\%$ macro-F1. 

\begin{table}[!htbp]
\caption{Ablation of response-target formatting on VAD-R. Ans-R and Abs-R denote answer recall and abstention recall, respectively.}
\label{tab:target_data_ablation}
\centering
\setlength{\tabcolsep}{2.2pt}
\scriptsize
\resizebox{0.62\linewidth}{!}{%
\begin{tabular}{llcccc}
\toprule
\textbf{Response Target} & \textbf{Loss} & \textbf{A-Acc} & \textbf{Ans-R} & \textbf{Abs-R} & \textbf{M-F1} \\
\midrule
Short w/o CoT & LM Only & 63.00 & \textbf{98.00} & 28.00 & 57.83 \\
Short w/ CoT & LM Only & 77.33 & 81.33 & \underline{73.33} & \underline{77.30} \\
Detailed w/ CoT & LM Only & 76.67 & 88.00 & 65.33 & 76.36 \\
Detailed w/o CoT & LM Only & \underline{77.67} & \underline{92.00} & 63.33 & 77.20 \\
\midrule
Rep2Act & Full objective & \textbf{86.33} & 86.67 & \textbf{86.00} & \textbf{86.33} \\
\bottomrule
\end{tabular}%
}
\end{table}

\subsection{Feature Anchoring versus Direct Routing Supervision}
\label{sec:appendix_centroid_geometry}

\paragraph{Feature Anchor Method.}
We implement this approach using an anchoring loss that minimizes the discrepancy between the student model’s direct-answer projection score and the frozen base model’s judgment-mode projection score through the Huber penalty $H$:
\begin{equation}
\label{eq:loss_anchor}
\mathcal L_{\mathrm{anchor}}
=\frac{1}{|\mathcal D_{\mathrm{train}}|}
\sum_{x\in\mathcal D_{\mathrm{train}}}
H\!\left(z_{D\to J,l}^{\theta}(x)-z_{J,l}^{\mathrm{base}}(x)\right),\quad
H(e)=\begin{cases}\tfrac12 e^2,&|e|\le1\\ |e|-\tfrac12,&|e|>1\end{cases}
\end{equation}
Here, $\theta$ denotes the trainable parameters of the student model.

The feature-anchor baseline uses layer $l=31$ of Qwen2.5-VL-3B-Instruct for both the student Direct features and the frozen Base Judge features. FA-Route-SFT trains the model to predict the correct A/U token and generate the reference response, as in Route-SFT. It also uses a Huber loss, weighted by 0.1, to bring the student's answerability projection under Direct prompting closer to the frozen base model's projection under Judge prompting.

\paragraph{Results and Analysis}
Table~\ref{tab:route_feature_ablation} shows that feature anchoring improves A-Acc by only $0.66\%$ over standard Route-SFT. To understand this limited gain, we analyze the cross-prompt centroid shift and the objective optimized by the anchoring loss. 
The centroids and judgment-mode answerability direction are estimated from the training representations. 
For the base Qwen2.5-VL-3B model at the final layer (hidden-state index 36), the parallel and orthogonal components have $\ell_2$ norms of $\lVert\Delta\vc_{\parallel}\rVert_2=10.33$ and $\lVert\Delta\vc_{\perp}\rVert_2=91.09$, respectively. 
The dominant orthogonal component may partly reflect the change in task induced by prompt switching: the model transitions from answering a question to determining whether that question is answerable. Prior work similarly demonstrates that prompts can alter the task encoded in hidden representations~\citep{hendel2023taskvectors,todd2024functionvectors,kirsanov2025geometry}. Plausible contributors to $\Delta\vc_{\perp}$ include:
\begin{itemize}[leftmargin=*]
\item the current task mode: directly answering a question or judging whether it is answerable;
\item the expected output space: an open-ended answer or a binary yes/no judgment;
\item the content and length of the instruction;
\item the identity and position of the final prompt token from which the representation is extracted;
\item prompt-dependent attention pathways;
\item the model’s internal preparation for the subsequent decoding task.
\end{itemize}
Together, these factors may account for the substantial orthogonal shift. 
Because the anchoring loss aligns only the models’ projection scores along the answerability direction, its influence on the final A/U token decision remains indirect.

\begin{figure}[!tb]
\centering
\includegraphics[width=0.92\textwidth]{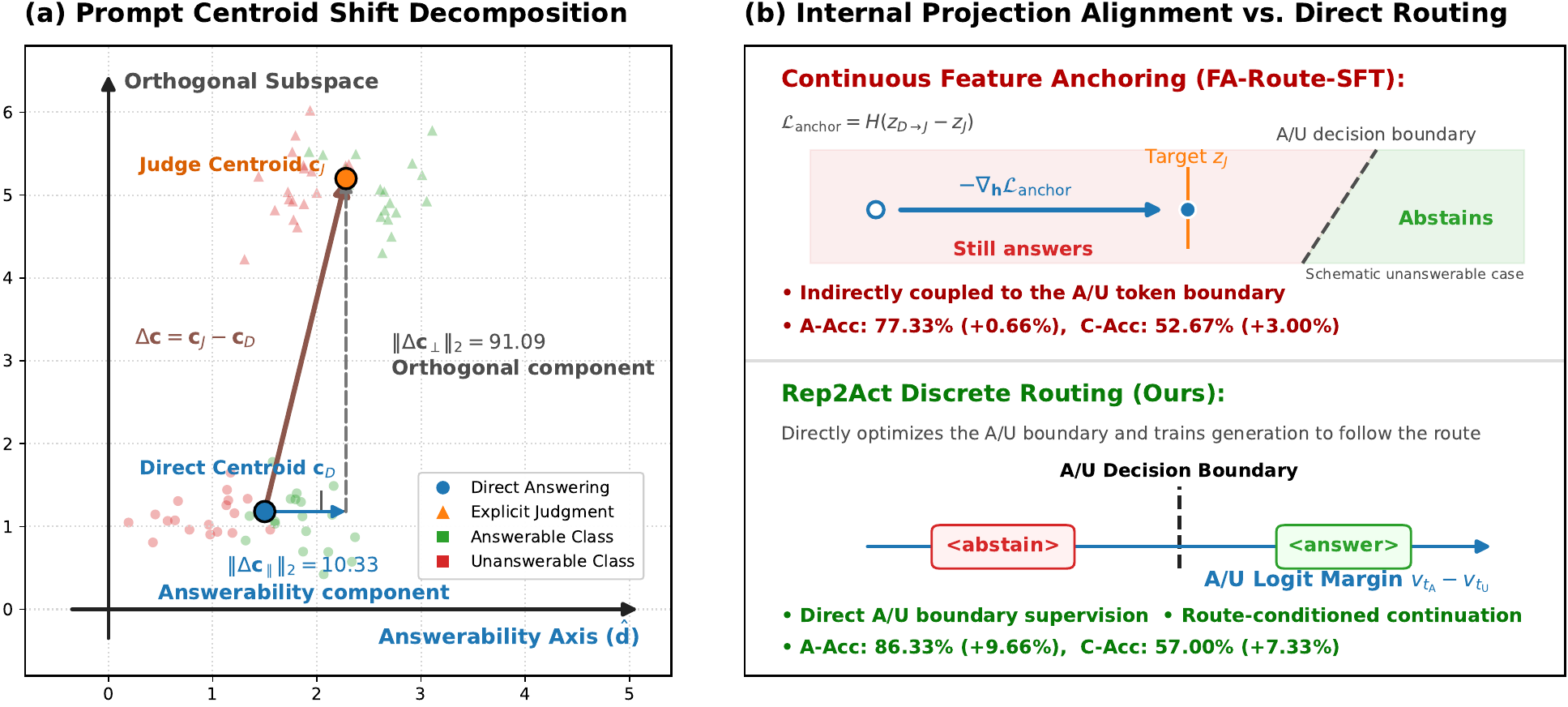}
\caption{\small Geometric decomposition of prompt centroid shift and comparison between internal projection alignment and direct routing supervision. \textbf{(a)}: The total displacement $\Delta\vc = \vc_J - \vc_D$ contains a component parallel to the answerability direction, $\Delta\vc_\parallel$, and a larger orthogonal prompt-mode component, $\Delta\vc_\perp$. \textbf{(b)}: Feature anchoring aligns the Direct cross-prompt score $z_{D\to J}$ with the frozen Judge score $z_J$, whereas Rep2Act directly optimizes the A/U decision boundary and conditions the continuation on the selected route.}
\label{fig:centroid_shift_geometry}
\end{figure}

% \paragraph{Gradient Updates.}
The anchor loss uses the student's Direct feature $\vh_l^{D,\theta}(x)$ and the frozen Base Judge feature $\vh_l^{J,\mathrm{base}}(x)$ at layer $l=31$. Both are projected onto the frozen unit Judge direction $\hat{\vd}_{J,l}^{\mathrm{base}}$. For $N$ training samples and prompt mode $M\in\{D,J\}$, the implemented centering and normalization are
\begin{equation}
\begin{aligned}
\vc_{M,l}^{\mathrm{base}}&=\frac{1}{N}\sum_{i=1}^{N}\vh_l^{M,\mathrm{base}}(x_i)\\
\sigma_{M\to J,l}^{\mathrm{base}}&=\sqrt{\frac{1}{N-1}\sum_{i=1}^{N}\left[(\vh_l^{M,\mathrm{base}}(x_i)-\vc_{M,l}^{\mathrm{base}})^\top\hat{\vd}_{J,l}^{\mathrm{base}}\right]^2}
\end{aligned}
\end{equation}
Thus, both scales measure variation along the Judge direction, with $\sigma_{J\to J,l}^{\mathrm{base}}=\sigma_{J,l}^{\mathrm{base}}$. This is the same sample-standard-deviation normalization used in the heatmaps. The centers, scales, direction, and Judge target are fixed during training. The two scores used by the anchor loss are
\begin{equation}
\label{eq:anchor_normalized_scores}
\begin{aligned}
z_{D\to J,l}^\theta(x)&=\frac{(\vh_l^{D,\theta}(x)-\vc_{D,l}^{\mathrm{base}})^\top\hat{\vd}_{J,l}^{\mathrm{base}}}{\sigma_{D\to J,l}^{\mathrm{base}}}\\
z_{J,l}^{\mathrm{base}}(x)&=\frac{(\vh_l^{J,\mathrm{base}}(x)-\vc_{J,l}^{\mathrm{base}})^\top\hat{\vd}_{J,l}^{\mathrm{base}}}{\sigma_{J,l}^{\mathrm{base}}}
\end{aligned}
\end{equation}
To differentiate the loss for one sample, write $\vh=\vh_l^{D,\theta}(x)$ and $e=z_{D\to J,l}^\theta(x)-z_{J,l}^{\mathrm{base}}(x)$. Since Judge target is frozen, its derivative with respect to $\vh$ is zero. Differentiating the linear projection gives
\begin{equation}
\nabla_{\vh}e
=\nabla_{\vh}z_{D\to J,l}^\theta(x)
=\frac{\hat{\vd}_{J,l}^{\mathrm{base}}}{\sigma_{D\to J,l}^{\mathrm{base}}}
\end{equation}
The implemented Huber loss has threshold 1: $H(e)=\tfrac12e^2$ for $|e|\le1$ and $H(e)=|e|-\tfrac12$ otherwise. Applying the chain rule yields
\begin{equation}
\label{eq:anchor_hidden_gradient}
\begin{aligned}
\nabla_{\vh}H(e)
&=\frac{\mathrm dH(e)}{\mathrm de}\,\nabla_{\vh}e\\
&=\begin{cases}
\displaystyle e\,\frac{\hat{\vd}_{J,l}^{\mathrm{base}}}{\sigma_{D\to J,l}^{\mathrm{base}}},& |e|\le1\\[4pt]
\displaystyle \operatorname{sign}(e)\,\frac{\hat{\vd}_{J,l}^{\mathrm{base}}}{\sigma_{D\to J,l}^{\mathrm{base}}},& |e|>1
\end{cases}
\end{aligned}
\end{equation}
For the weighted mean anchor loss over $n$ eligible VAD-R samples, the contribution of sample $i$ to its hidden-state gradient is $\frac{0.1}{n}\nabla_{\vh_i}H(e_i)$, using the implemented anchor weight 0.1. The parameter gradient follows by backpropagating through the student feature:
\begin{equation}
\nabla_\theta\bigl(0.1\mathcal L_{\mathrm{anchor}}\bigr)
=\frac{0.1}{n}\sum_{i=1}^{n}
\left(\frac{\partial\vh_i}{\partial\theta}\right)^\top
\nabla_{\vh_i}H(e_i) 
\end{equation}
Here $\operatorname{sign}(e)$ is $+1$ for positive errors and $-1$ for negative errors. 

Equation~\ref{eq:anchor_hidden_gradient} shows that the gradient with respect to the hidden representation is a scalar multiple of the fixed judgment-mode direction. When the student’s projection is too high ($e>0$), gradient descent decreases it; when the projection is too low ($e<0$), gradient descent increases it. The anchoring loss therefore trains the student to match the judgment-mode projection. However, because this gradient depends on the projection error rather than the correctness of the A/U token decision, it influences that decision only indirectly. In contrast, Rep2Act’s signed margin loss derives its gradient directly from the ground-truth A/U label and the competing token logits, while the route-contrastive loss further encourages the generated response to remain consistent with the correct route.

\subsection{Sensitivity Analysis of Training-Loss Hyperparameters}
\label{sec:appendix_loss_sensitivity}
Table~\ref{tab:appendix_loss_sensitivity} presents detailed sensitivity analyses of the signed margin loss weight $\lambda_m$ and route-contrastive loss weight $\lambda_c$ on VAD-R. 
% Performance under both loss components follows a characteristic inverted-U pattern: 

\paragraph{Signed Margin Loss Weight $\lambda_m$.}  Signed Margin Loss directly widens the signed vocabulary-logit difference $s(v_{t_{\mathrm A}}-v_{t_{\mathrm U}})$ at the A/U decision step. At $\lambda_m = 0.0$, the model reaches 74.33\% A-Acc. Increasing $\lambda_m$ to 0.2 raises accuracy to 82.33\% and yields balanced recall between answer and abstain categories, at 81.33\% and 83.33\% respectively. Further increasing $\lambda_m$ to 0.5 reduces accuracy to 80.67\% as the model begins favoring the answer route. Raising $\lambda_m$ to 1.0 and 2.0 decreases abstain recall to 66.67\% and 61.33\%, bringing final accuracy down to 75.67\%.

\paragraph{Route-Contrastive Loss Weight $\lambda_c$.} Route-Contrastive Loss penalizes counterfactual generations, ensuring that the gold continuation remains significantly more probable under the chosen route than under the inverted route. At $\lambda_c = 0.0$, route-action consistency is 89.00\% and A-Acc is 78.33\%. Increasing $\lambda_c$ through 0.1, 0.5, and 1.0 progressively elevates A-Acc to 82.33\%, 85.00\%, and 86.33\%, while driving consistency up to 93.67\%, 99.00\%, and 99.33\%. Setting $\lambda_c$ to 2.0 slightly degrades A-Acc to 85.33\%  at this larger loss weight.

\begin{table*}[!htb]
\caption{Hyperparameter sensitivity analysis on VAD-R. Cons.\ denotes the percentage of route–action-consistent responses.} 
\label{tab:appendix_loss_sensitivity}
\centering
\begin{subtable}[t]{0.485\textwidth}
\caption{Margin Loss weight $\lambda_m$, with fixed $\lambda_c=0.1$.}
\label{tab:appendix_margin_sensitivity}
\centering
\setlength{\tabcolsep}{4.2pt}
\scriptsize
\resizebox{\linewidth}{!}{%
\begin{tabular}{cccccc}
\toprule
\textbf{$\lambda_m$} & \textbf{A-Acc} & \textbf{M-P} & \textbf{M-R} & \textbf{M-F1} & \textbf{Cons.} \\
\midrule
0.0 & 74.33 & 74.39 & 74.33 & 74.32 & \textbf{98.67} \\
0.2 & \textbf{82.33} & \textbf{82.35} & \textbf{82.33} & \textbf{82.33} & \underline{93.67} \\
0.5 & \underline{80.67} & \underline{81.47} & \underline{80.67} & \underline{80.54} & 89.33 \\
1.0 & 77.67 & 79.07 & 77.67 & 77.39 & 84.33 \\
2.0 & 75.67 & 77.96 & 75.67 & 75.16 & 80.67 \\
\bottomrule
\end{tabular}%
}
\end{subtable}%
\hfill
\begin{subtable}[t]{0.485\textwidth}
\caption{RC Loss weight $\lambda_c$, with fixed $\lambda_m=0.2$.}
\label{tab:appendix_contrastive_sensitivity}
\centering
\setlength{\tabcolsep}{4.2pt}
\scriptsize
\resizebox{\linewidth}{!}{%
\begin{tabular}{cccccc}
\toprule
\textbf{$\lambda_c$} & \textbf{A-Acc} & \textbf{M-P} & \textbf{M-R} & \textbf{M-F1} & \textbf{Cons.} \\
\midrule
0.0 & 78.33 & 78.55 & 78.33 & 78.29 & 89.00 \\
0.1 & 82.33 & 82.35 & 82.33 & 82.33 & 93.67 \\
0.5 & 85.00 & 85.26 & 85.00 & 84.97 & \underline{99.00} \\
1.0 & \textbf{86.33} & \textbf{86.33} & \textbf{86.33} & \textbf{86.33} & \textbf{99.33} \\
2.0 & \underline{85.33} & \underline{85.34} & \underline{85.33} & \underline{85.33} & 98.33 \\
\bottomrule
\end{tabular}%
}
\end{subtable}
\end{table*}
\clearpage
\section{Prompt Templates}
\label{sec:appendix_prompts}

This section documents the prompt templates used throughout our experiments, including those for baseline methods, diagnostic probing, and automated evaluation by the action and content judges. 

\subsection{Baselines and Probing Prompts}
\label{sec:appendix_prompts_baselines}

\noindent\textbf{Direct Answering ($P_D$)} requests a response without explicitly eliciting intermediate reasoning: 

\begin{tcolorbox}[colback=gray!4!white,colframe=gray!60!black,title=\textbf{Prompt Template: Direct Answering ($P_D$)},fonttitle=\bfseries\footnotesize,width=\textwidth,top=0.8mm,bottom=0.8mm,boxsep=0.8mm,before=\vspace{0.5mm},after=\vspace{0.5mm}]
\footnotesize\ttfamily
\{question\}
\end{tcolorbox}

\noindent\textbf{Explicit Judgment ($P_J$)} probes whether the base model can explicitly assess answerability through a binary query: 

\begin{tcolorbox}[colback=gray!4!white,colframe=gray!60!black,title=\textbf{Prompt Template: Explicit Judgment ($P_J$)},fonttitle=\bfseries\footnotesize,width=\textwidth,top=0.8mm,bottom=0.8mm,boxsep=0.8mm,before=\vspace{0.5mm},after=\vspace{0.5mm}]
\footnotesize\ttfamily
Based only on the image, can a viewer reliably answer the question?\\
Answer only yes or no.\\[1mm]
Question: \{question\}
\end{tcolorbox}

\noindent\textbf{Judge-then-Answer ($P_{JA}$)} explicitly separates the assessment of visual-evidence sufficiency from subsequent answer generation or the articulation of an abstention rationale:

\begin{tcolorbox}[colback=gray!4!white,colframe=gray!60!black,title=\textbf{Prompt Template: Judge-then-Answer ($P_{JA}$)},fonttitle=\bfseries\footnotesize,width=\textwidth,top=0.8mm,bottom=0.8mm,boxsep=0.8mm,before=\vspace{0.5mm},after=\vspace{0.5mm}]
\footnotesize\ttfamily
You are given an image and a question.\\
Please follow these two steps:\\
Step 1 [Judgment]: First determine whether the visual evidence in the image is sufficient to answer the question.\\
- If the question cannot be answered from the image (e.g., the required object or information is missing, occluded, unanswerable, or ambiguous), output: ``Judgment: Unanswerable'' and explain why.\\
- If the question can be answered from the image, output: ``Judgment: Answerable'', followed by the concise answer in the format ``Answer: <your answer>''.\\[1mm]
Question: \{question\}
\end{tcolorbox}

\subsection{Automatic Judge Prompts}
\label{sec:appendix_prompts_judges}

\noindent\textbf{Action Judge: Qwen3.6-27B} classifies each response as either an attempted answer or evidence-grounded abstention, using greedy decoding with reasoning disabled: 

\begin{tcolorbox}[colback=gray!4!white,colframe=gray!60!black,title=\textbf{Prompt Template: Action Judge: Strict Binary Action Classifier},fonttitle=\bfseries\footnotesize,width=\textwidth,top=0.8mm,bottom=0.8mm,boxsep=0.8mm,before=\vspace{0.5mm},after=\vspace{0.5mm}]
\footnotesize\ttfamily
\textbf{[System Prompt]}\\
You are a strict binary classifier. Do not reveal reasoning.\\[1mm]
\textbf{[User Prompt]}\\
Question:\\
\{question\}\\[1mm]
Model response:\\
\{response\}\\[1mm]
Does the model response refuse to answer because the image does not provide enough evidence?\\
Do not explain. End with exactly one word: yes or no.
\end{tcolorbox}

\noindent\textbf{Content Judge: Qwen3.8-27B} evaluates C-Acc on VAD-R using greedy decoding with reasoning disabled. For answerable queries ($y=\mathrm{A}$), it assesses the factual correctness of each response against the visual evidence and reference answer: 

\begin{tcolorbox}[colback=gray!4!white,colframe=gray!60!black,title=\textbf{Prompt Template: Content Judge: Answerable Verification ($y=\text{A}$)},fonttitle=\bfseries\footnotesize,width=\textwidth,top=0.8mm,bottom=0.8mm,boxsep=0.8mm,before=\vspace{0.5mm},after=\vspace{0.5mm}]
\footnotesize\ttfamily
\textbf{[User Prompt]}\\
Question:\\
\{question\}\\[1mm]
Reference Answer:\\
\{reference\_answer\}\\[1mm]
Candidate Response:\\
\{candidate\_response\}\\[1mm]
Based on the image, is the candidate response correct? Reply with exactly yes or no.
\end{tcolorbox}

\noindent For unanswerable queries ($y=\mathrm{U}$), Qwen3.8-27B determines whether the candidate response provides a valid explanation that correctly identifies the missing visual evidence: 

\begin{tcolorbox}[colback=gray!4!white,colframe=gray!60!black,title=\textbf{Prompt Template: Content Judge: Evidence-Gap Explanation Verification ($y=\text{U}$)},fonttitle=\bfseries\footnotesize,width=\textwidth,top=0.8mm,bottom=0.8mm,boxsep=0.8mm,before=\vspace{0.5mm},after=\vspace{0.5mm}]
\footnotesize\ttfamily
\textbf{[User Prompt]}\\
Question:\\
\{question\}\\[1mm]
Reference Missing-Evidence Explanation:\\
\{reference\_reason\}\\[1mm]
Candidate Response:\\
\{candidate\_response\}\\[1mm]
Based on the image, does the candidate response explicitly state the correct missing visual evidence that makes the question impossible to answer reliably? A refusal without a reason is no. Reply with exactly yes or no.
\end{tcolorbox}

\WFclear\newpage
\section{Additional Response Examples}
\label{sec:additional_response_cases}
\begin{figure}[htbp]
\centering
\includegraphics[width=\textwidth]{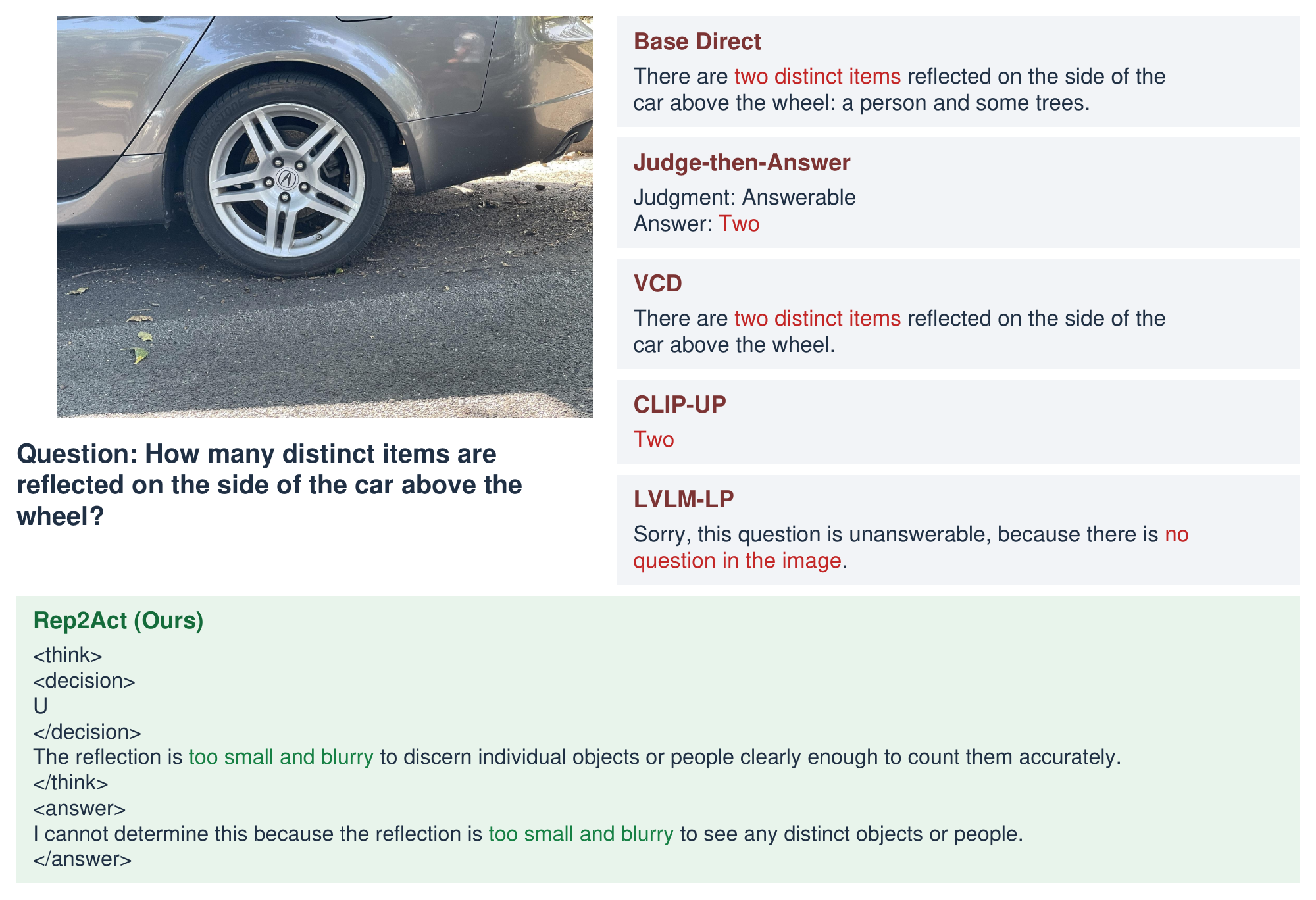}
\caption{VAD-R counting example. The baselines either provide a definitive count or abstain for an unrelated reason when asked about objects reflected on the car. \sys{} correctly identifies that the reflection is too indistinct to support a reliable count. Green and gray denote responses that pass and fail Qwen3.8-27B content judge evaluation, respectively.}
\label{fig:vadr_reflection_case}
\end{figure}

\begin{figure}[htbp]
\centering
\includegraphics[width=\textwidth]{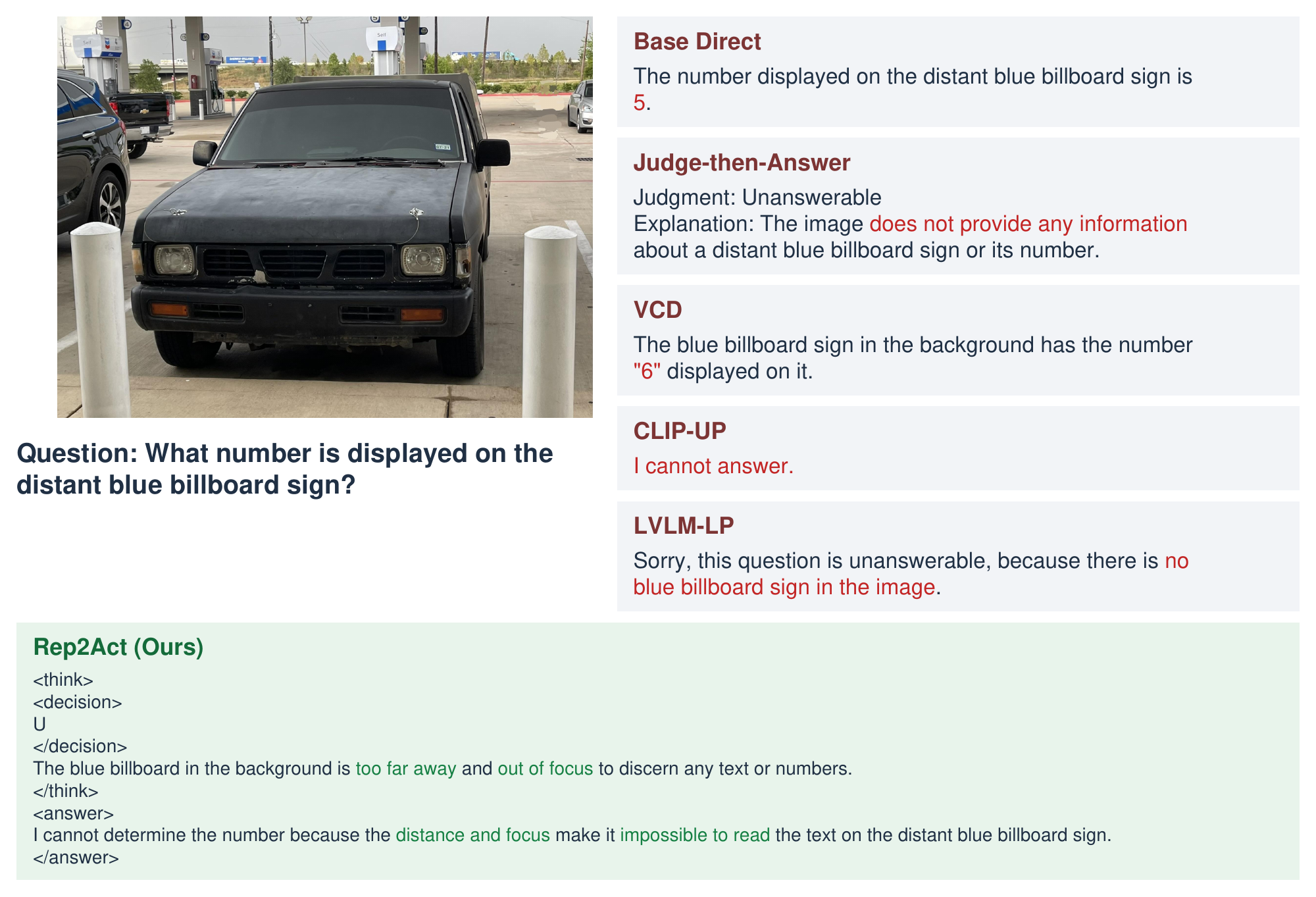}
\caption{VAD-R text-reading example. \emph{Base Direct} and \emph{VCD} produce inconsistent numbers for a distant billboard, while \emph{Judge-then-Answer} and \emph{LVLM-LP} incorrectly claim that the billboard is absent. \sys{} correctly explains that the text is unreadable because of the viewing distance and image focus. Green and gray denote responses that pass and fail Qwen3.8-27B content judge evaluation, respectively.}
\label{fig:vadr_billboard_case}
\end{figure}

Figures~\ref{fig:vadr_reflection_case} and~\ref{fig:vadr_billboard_case} present two additional response comparisons on VAD-R across 6 methods shown in Figure~\ref{fig:rep2act_algorithm_overview}(d), all built on the same 3B base model. 

\WFclear\newpage

\end{document}